\documentclass[10pt,twocolumn,letterpaper]{article}

\usepackage{cvpr}              

\usepackage{graphicx}
\usepackage{amsmath}
\usepackage{amssymb}
\usepackage{booktabs}

\usepackage{multirow}
\usepackage{verbatim}
\usepackage{enumitem}
\usepackage{bm}

\usepackage{xcolor}
\definecolor{bittersweet}{rgb}{1.0, 0.44, 0.37}
\definecolor{mygreen}{rgb}{0.29, 0.7, 0.48}

\usepackage[font=small,labelfont=bf]{caption}  
\usepackage{makecell}
\usepackage{tabulary}
\definecolor{demphcolor}{RGB}{144,144,144}
\newcommand{\demph}[1]{\textcolor{demphcolor}{#1}}
\definecolor{mygray}{gray}{0.4}
\usepackage{pifont}
\newlength\savewidth\newcommand\shline{\noalign{\global\savewidth\arrayrulewidth
  \global\arrayrulewidth 1pt}\hline\noalign{\global\arrayrulewidth\savewidth}}

\newcommand{\tablestyle}[2]{\setlength{\tabcolsep}{#1}\renewcommand{\arraystretch}{#2}\centering\footnotesize}
\makeatletter\renewcommand\paragraph{\@startsection{paragraph}{4}{\z@}
  {.5em \@plus1ex \@minus.2ex}{-.5em}{\normalfont\normalsize\bfseries}}\makeatother
\usepackage{xspace}
\def\ModelName{\textsc{Meter}\xspace}

\newcolumntype{C}[1]{>{\centering\arraybackslash}p{#1}}
\newcolumntype{R}[1]{>{\raggedleft\arraybackslash}p{#1}}
\newcolumntype{L}[1]{>{\raggedright\arraybackslash}p{#1}}
\usepackage[normalem]{ulem}
\usepackage[pagebackref,breaklinks,colorlinks]{hyperref}

\usepackage[capitalize]{cleveref}
\crefname{section}{Sec.}{Secs.}
\Crefname{section}{Section}{Sections}
\Crefname{table}{Table}{Tables}
\crefname{table}{Tab.}{Tabs.}

\def\cvprPaperID{7782} 
\def\confName{CVPR}
\def\confYear{2022}

\usepackage{arydshln}

\begin{document}

\title{An Empirical Study of Training End-to-End Vision-and-Language Transformers}

\author{Zi-Yi Dou$^1$\thanks{\, Work was done when the author interned at Microsoft.}, Yichong Xu$^2$, Zhe Gan$^2$, Jianfeng Wang$^2$, Shuohang Wang$^2$, Lijuan Wang$^2$,  \\
Chenguang Zhu$^2$, Pengchuan Zhang$^2$, Lu Yuan$^2$, Nanyun Peng$^1$, Zicheng Liu$^2$, Michael Zeng$^2$ \\
$^1$University of California, Los Angeles, $^2$Microsoft Corporation\\
{\tt\small \{zdou,violetpeng\}@cs.ucla.edu} \\
{\tt\small \{yicxu,zhgan,jianfw,shuowa,lijuanw,chezhu,penzhan,luyuan,zliu,nzeng\}@microsoft.com}
}
\maketitle

\begin{abstract}
   Vision-and-language (VL) pre-training has proven to be highly effective on various VL downstream tasks. While recent work has shown that fully transformer-based VL models can be more efficient than previous region-feature-based methods, their performance on downstream tasks often degrades significantly. In this paper, we present \ModelName, a \textbf{M}ultimodal \textbf{E}nd-to-end \textbf{T}ransform\textbf{ER} framework, through which we investigate how to design and pre-train a fully transformer-based VL model in an end-to-end manner. Specifically, we dissect the model designs along multiple dimensions: vision encoders (e.g., CLIP-ViT, Swin transformer), text encoders (e.g., RoBERTa, DeBERTa), multimodal fusion module (e.g., merged attention vs. co-attention), architectural design (e.g., encoder-only vs. encoder-decoder), and pre-training objectives (e.g., masked image modeling). We conduct comprehensive experiments and provide insights on how to train a performant VL transformer. \ModelName achieves an accuracy of 77.64\% on the VQAv2 test-std set using only 4M images for pre-training, surpassing the state-of-the-art region-feature-based model by 1.04\%, and outperforming the previous best 
   fully transformer-based model by 1.6\%. Notably, when further scaled up, our best VQA model achieves an accuracy of 80.54\%. Code and pre-trained models are released at \url{https://github.com/zdou0830/METER}.
\end{abstract}
 \vspace{-5mm}
\section{Introduction}
\label{sec:intro}
Vision-and-language (VL) tasks, such as visual question answering (VQA)~\cite{antol2015vqa} and image-text retrieval~\cite{lin2014microsoft,plummer2015flickr30k}, require an AI system to comprehend both the input image and text contents. Vision-and-language pre-training (VLP) has now become the \emph{de facto} practice to tackle these tasks~\cite{tan-bansal-2019-lxmert,li2019visualbert,lu2019vilbert,su2019vl,chen2020uniter,li2020oscar}.
Specifically, large amounts of image-caption pairs are fed into a model that consumes both images and texts to pretrain representations that contain rich multimodal knowledge and is helpful for downstream tasks.

\begin{figure}[t]
  \centering
   \includegraphics[width=1.0\linewidth]{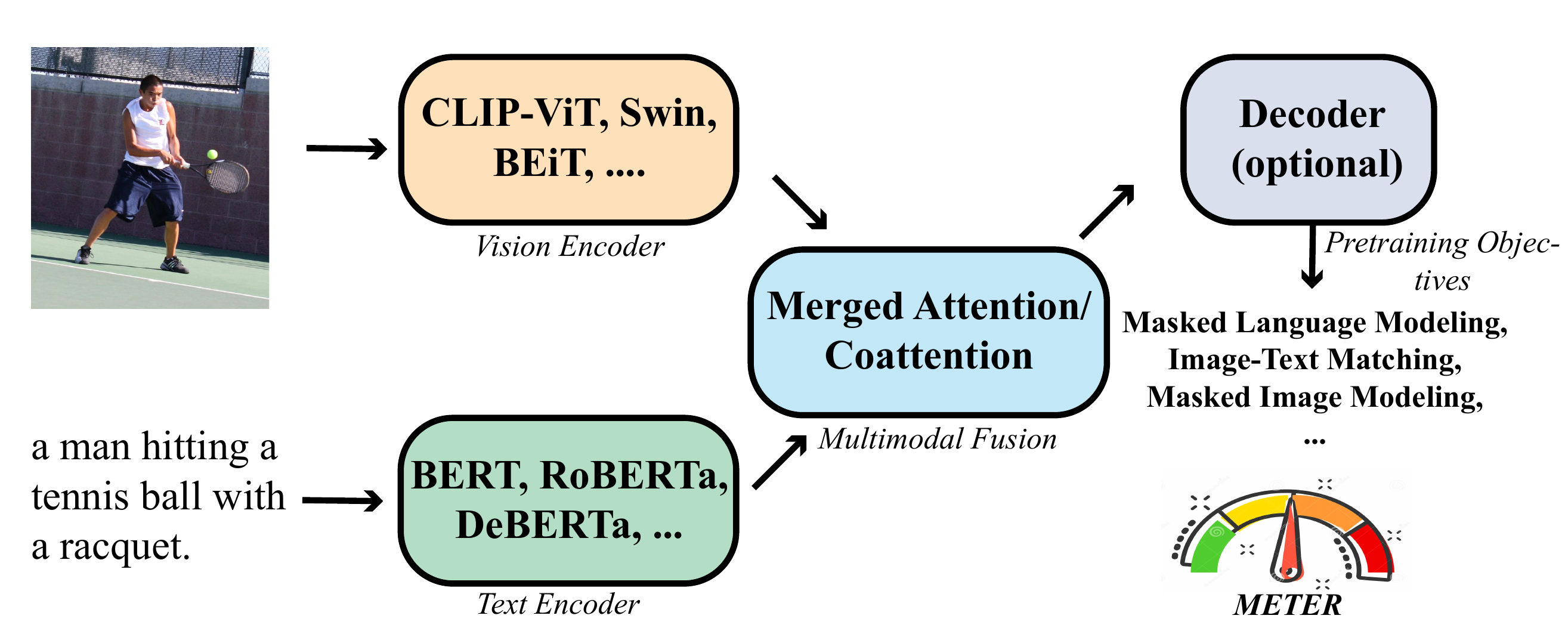}
   \caption{\textbf{An overview of the proposed \ModelName~framework}.
   We systematically investigate how to train a performant vision-and-language transformer, and dissect the model designs along multiple dimensions: vision encoder, text encoder, multimodal fusion module, architectural design (encoder-only vs. encoder-decoder), and pre-training objectives.}
   \label{fig:overview_meter}
   \vspace{-5mm}
\end{figure}

\begin{table*}[!t]
\tablestyle{5pt}{1.0}
\def\w{20pt} 
\scalebox{1.0}{
  \begin{tabular}{l|cccccccccccccccccccccccc}
    \bf Model & \bf Vision Encoder & \bf Text Encoder  & \bf Multimodal Fusion & \bf Decoder & \bf Pre-training Objectives \\
    \shline
    ViLBERT~\cite{lu2019vilbert} & 
\multirow{2}{*}{OD+Xformer}  & \multirow{2}{*}{Xformer}    & \multirow{2}{*}{Co-attn.} &  \multirow{7}{*}{\ding{55} } & MLM+ITM+MIM\\
LXMERT~\cite{tan-bansal-2019-lxmert} & &   & &   & MLM+ITM+MIM+VQA \\
    \cdashline{2-4}
    VisualBERT~\cite{li2019visualbert} & \multirow{6}{*}{OD}   & \multirow{6}{*}{Emb.}  & \multirow{6}{*}{Merged-attn.} &    & MLM+ITM\\
 VL-BERT~\cite{su2019vl} &  &  & &    & MLM+MIM\\
  UNITER~\cite{chen2020uniter} & &   & & & MLM+ITM+MIM+WRA \\
  OSCAR~\cite{li2020oscar} & &   & &  & MLM+ITM \\
  VinVL~\cite{zhang2021vinvl} & &  & &  & MLM+ITM\\
  \cdashline{5-5}
   VL-T5~\cite{cho2021unifying}  &   &    & & \ding{51} & MLM+ITM+VQA+Grounding+Captioning\\
  \hline
  PixelBERT~\cite{huang2020pixel} & \multirow{5}{*}{CNN} & \multirow{4}{*}{Emb.}  & \multirow{5}{*}{Merged-attn.} & \multirow{3}{*}{\ding{55} } & MLM+ITM\\
  SOHO~\cite{huang2021seeing} & &  & &  & MLM+ITM+MIM\\
  CLIP-ViL~\cite{shen2021much}  & & & &  & MLM+ITM+VQA\\
  \cdashline{5-5}
   SimVLM~\cite{wang2021simvlm} &   &   &    & \multirow{2}{*}{\ding{51}} & PrefixLM\\
     \cdashline{3-3}
   MDETR~\cite{kamath2021mdetr} &   & Xformer &  & & OD+Token Prediction+Contrastive Alignment\\
  \hline
  ViLT~\cite{kim2021vilt}&  \multirow{1}{*}{Patch Emb.} & \multirow{2}{*}{Emb.}  & \multirow{2}{*}{Merged-attn.} & \multirow{4}{*}{\ding{55}} & MLM+ITM\\
   \cdashline{2-2}
  Visual Parsing~\cite{xue2021probing} & \multirow{3}{*}{Xformer} &  & &   & MLM+ITM+MIM\\
   \cdashline{3-4}
  ALBEF~\cite{li2021align} &   &   \multirow{2}{*}{Xformer}  & \multirow{2}{*}{Co-attn.}&   & MLM+ITM+ITC\\
  \ModelName~(Ours) &  &    &  &  & MLM+ITM\\
  \hline
  CLIP~\cite{radford2021learning}  & CNN/Xformer & \multirow{2}{*}{Xformer} & \multirow{2}{*}{None} &\multirow{2}{*}{\ding{55}}  & \multirow{2}{*}{ITC}\\
  ALIGN~\cite{jia2021scaling}  & CNN & &  &   & \\
  \end{tabular}
  }
  \caption{\textbf{Glossary of representative VLP models}. OD: objective detector. Xformer: transformer. Emb.: embedding. MLM/MIM: masked language/image modeling. ITM: image-text matching. WRA: word-region alginment. ITC: image-text contrastive learning.}
  \label{tab:vlp_glossary}
  \vspace{-4mm}
\end{table*}

Transformers~\cite{vaswani2017attention} are prevalent in natural language processing and have recently shown promising performance in computer vision~\cite{dosovitskiy2020image,liu2021swin}. While almost all the existing VLP models adopt transformers for their multimodal fusion module, most of them~\cite{tan-bansal-2019-lxmert,li2019visualbert,lu2019vilbert,su2019vl,chen2020uniter,li2020oscar} use pre-trained object detectors (\emph{e.g.}, Faster RCNN~\cite{ren2015faster}) on the vision side to extract \emph{region} features from images. This can lead to several problems: first, the object detectors are not perfect, but are usually kept frozen during VLP, which limits the capacity of the VLP models; second, it is time-consuming to extract region features~\cite{kim2021vilt}. 
On the other hand, vision transformers (ViTs) have been an increasingly active research topic in computer vision and have shown great potential in vision feature extraction. 
Therefore, a natural question arises:
\emph{can we train a fully transformer-based VLP model with ViTs as the image encoder?} 

Recent works~\cite{kim2021vilt,xue2021probing,li2021align} that tried to adopt vision transformers have not shown satisfactory performance and typically underperform state-of-the-art region-feature-based VLP models (\emph{e.g.}, VinVL~\cite{zhang2021vinvl}). To close the performance gap, we present \ModelName, a \textbf{M}ultimodal \textbf{E}nd-to-end \textbf{T}ransform\textbf{ER} framework, through which we thoroughly investigate how to design and pre-train a fully transformer-based VLP model in an end-to-end manner. Specifically, as shown in Figure~\ref{fig:overview_meter}, we dissect the model designs along multiple dimensions, including vision encoders (\emph{e.g.}, CLIP-ViT~\cite{radford2021learning}, Swin transformer~\cite{liu2021swin}), text encoders (\emph{e.g.}, RoBERTa~\cite{liu2019roberta}, DeBERTa~\cite{he2020deberta}), multimodal fusion module (\emph{e.g.}, merged attention vs. co-attention), architectural design (\emph{e.g.}, encoder-only vs. encoder-decoder), and pre-training objectives (\emph{e.g.}, masked image modeling~\cite{bao2021beit}).


We perform the investigation by pre-training models under \ModelName on four commonly used image-caption datasets: COCO~\cite{lin2014microsoft}, Conceptual Captions~\cite{sharma2018conceptual}, SBU Captions~\cite{ordonez2011im2text}, and Visual Genome~\cite{krishna2016visual}. We test them on visual question answering~\cite{antol2015vqa}, visual reasoning~\cite{suhr2018corpus}, image-text retrieval~\cite{lin2014microsoft,plummer2015flickr30k}, and visual entailment~\cite{xie2019visual} tasks. Through extensive analyses, we summarize our findings as follows:
\begin{itemize}[leftmargin=*]
    \item Vision transformer (ViT) plays a more vital role than language transformer in VLP, and the performance of transformers on pure vision or language tasks is not a good indicator for its performance on VL tasks.
    \vspace{-2mm}
    \item The inclusion of cross-attention benefits multimodal fusion, which results in better downstream performance than using self-attention alone.  
    \vspace{-2mm}
    \item Under a fair comparison setup, the encoder-only VLP model performs better than the encoder-decoder model for VQA and zero-shot image-text retrieval tasks. 
    \vspace{-2mm}
    \item Adding the masked image modeling loss in VLP will \emph{not} improve downstream task performance in our settings.
\end{itemize}
These insights, combined with other useful tips and tricks detailed in later sections, enable us to train a strong model that achieves an accuracy of 77.64\% on the VQAv2 test-std set, surpassing the previous best region-feature-based VinVL model~\cite{zhang2021vinvl} by 1.04\% and outperforming the previously best ViT-based model (\ie, ALBEF~\cite{li2021align}) by 1.6\%. Notably, when further scaled up, our best \ModelName model achieves an accuracy of 80.54\% on the VQAv2 test-std set.


\section{Glossary of VLP Models}
In this section, we provide an overview of representative VLP models, and divide them into three categories based on how they encode images, as summarized in~Table \ref{tab:vlp_glossary}.
\paragraph{OD-based Region Features.} 
Most previous work use pre-trained object detectors (ODs) to extract visual features. Among them, ViLBERT~\cite{lu2019vilbert} and LXMERT~\cite{tan-bansal-2019-lxmert} use
co-attention for multimodal fusion, where two transformers are applied independently to region and text features, and another transformer fuses the representations of the two modalities in a later stage. On the other hand, VisualBERT~\cite{li2019visualbert}, 
VL-BERT~\cite{su2019vl}, and UNITER~\cite{chen2020uniter}
use a merged attention fusion module that feeds both region and text features together into a single transformer. OSCAR~\cite{li2020oscar} and VinVL~\cite{zhang2021vinvl} feed additional image tags into the transformer model, and demonstrate state-of-the-art performance across VL tasks. However, extracting region features can be time-consuming, and the pre-trained ODs are usually frozen during pre-training, which limits the capacity of VLP models.

\paragraph{CNN-based Grid Features.} To tackle the above two issues, researchers have tried different ways to pre-train VL models in an end-to-end fashion. Among them, PixelBERT~\cite{huang2020pixel} and CLIP-ViL~\cite{shen2021much} propose to feed grid features from convolutional neural networks (CNNs) and text directly into a transformer. SOHO~\cite{huang2021seeing} proposes to to first discretize grid features using a learned vision dictionary, then feed the discretized features into their cross-modal module. While using grid features directly can be efficient, inconsistent optimizers are typically used for CNN and transformer. For example, PixelBERT~\cite{huang2020pixel} and CLIP-ViL~\cite{shen2021much} use AdamW~\cite{loshchilov2018decoupled} for transformer and SGD for CNN. Recent work on vision transformers (ViTs) has also shown that CNN can achieve slightly worse accuracy/FLOPs trade-offs than their ViT counterparts~\cite{liu2021swin}, motivating researchers to develop ViT-based VLP models.

\paragraph{ViT-based Patch Features.}  ViLT~\cite{kim2021vilt} directly feeds image patch features and text token embeddings into a pre-trained ViT model, and fine-tunes the model on image-caption datasets. More recently, visual parsing\cite{xue2021probing} and ALBEF~\cite{li2021align} use ViT as the image encoder and design different pre-training objectives for ViT-based VLP models. However, all these models lag behind the state-of-the-art performance on downstream tasks such as visual question answering. In this paper, we investigate how to pre-train a ViT-based model in an end-to-end manner that closes the performance gap while maintaining fast inference speed.


\section{The \ModelName~Framework}
\label{sec:base}

Based on the previous work, we identify several important modules in VLP models as in Figure~\ref{fig:overview_meter}. In this section, we first illustrate our \ModelName~framework, then our default settings, which paves the way for our analyses hereinafter. 


\paragraph{Overview.} 
Given a text sentence $\mathbf{l} $ and an image $\mathbf{v}$, a VLP model first extracts both text features $\mathbf{l}=\langle l_1, \cdots, l_N\rangle$ and visual features $\mathbf{v}=\langle v_1, \cdots, v_M\rangle$ via a \textit{text encoder} and a \textit{vision encoder}. 
The text and visual features are then fed into a \textit{multimodal fusion module} to produce cross-modal representations, which are then optionally fed into a \textit{decoder} 
before generating the final outputs. 



\subsection{Model Architecture}
\paragraph{Vision Encoder.} 
In this paper, we focus on patch features, and study the use of vision transformers (ViTs)~\cite{dosovitskiy2020image} for vision encoder. Specifically, in ViT, an image is first segmented into patches, and then the patches are fed into a transformer model. ViT has become a popular research topic recently~\cite{dosovitskiy2020image,touvron2020deit,touvron2020deit,touvron2021going,yuan2021volo,liu2021swin,bao2021beit}, and has been introduced into VLP~\cite{kim2021vilt,xue2021probing,li2021align}. However, all these models only achieve inferior performance compared to state-of-the-art region-feature-based models (\emph{e.g.}, VinVL~\cite{zhang2021vinvl}). Also, different pre-trained ViTs are used, lacking a systematic study of which ViTs are the best for VLP.
In this work, we compare the original ViT~\cite{dosovitskiy2020image}, DeiT~\cite{touvron2020deit}, Distilled-DeiT~\cite{touvron2020deit}, CaiT~\cite{touvron2021going}, VOLO~\cite{yuan2021volo}, BEiT~\cite{bao2021beit}, Swin Transformer~\cite{liu2021swin} and CLIP-ViT~\cite{radford2021learning}, to provide a comprehensive analysis on the role of vision transformers.

\paragraph{Text Encoder.} Following BERT~\cite{devlin2018bert} and RoBERTa~\cite{liu2019roberta}, VLP models~\cite{tan-bansal-2019-lxmert,li2019visualbert,lu2019vilbert,su2019vl,chen2020uniter,li2020oscar} first segment the input sentence into a sequence of subwords~\cite{sennrich2016neural}, then insert two special tokens at the beginning and the end of the sentence to generate the input text sequence. After we obtain the text embeddings, existing works either feed them directly to the multimodal fusion module~\cite{li2019visualbert,chen2020uniter}, or to several text-specific layers~\cite{tan-bansal-2019-lxmert,lu2019vilbert} before the fusion. For the former, the fusion module is typically initialized with BERT, and the role of text encoding and multimodal fusion is therefore  entangled and absorbed in a single BERT model. Here, we aim to disentangle the two modules, and use a text encoder first before sending the features into the fusion module. 

Language model (LM) pre-training has demonstrated impressive performance across tasks and different pre-trained LMs have been proposed; however, most VLP models still only use BERT for initialization~\cite{chen2020uniter}. In this work, we study the use of BERT~\cite{devlin2018bert}, RoBERTa~\cite{liu2019roberta}, ELECTRA~\cite{clark2020electra}, ALBERT~\cite{lan2019albert}, and DeBERTa~\cite{he2020deberta} for text encoding. Besides, we also experiment on only using a simple word embedding look-up layer initialized with the BERT embedding layer as used in many previous works~\cite{chen2020uniter,zhang2021vinvl}. 

\begin{figure}[t!]
  \centering
  \begin{subfigure}{0.49\linewidth}
    \includegraphics[width=1.0\linewidth]{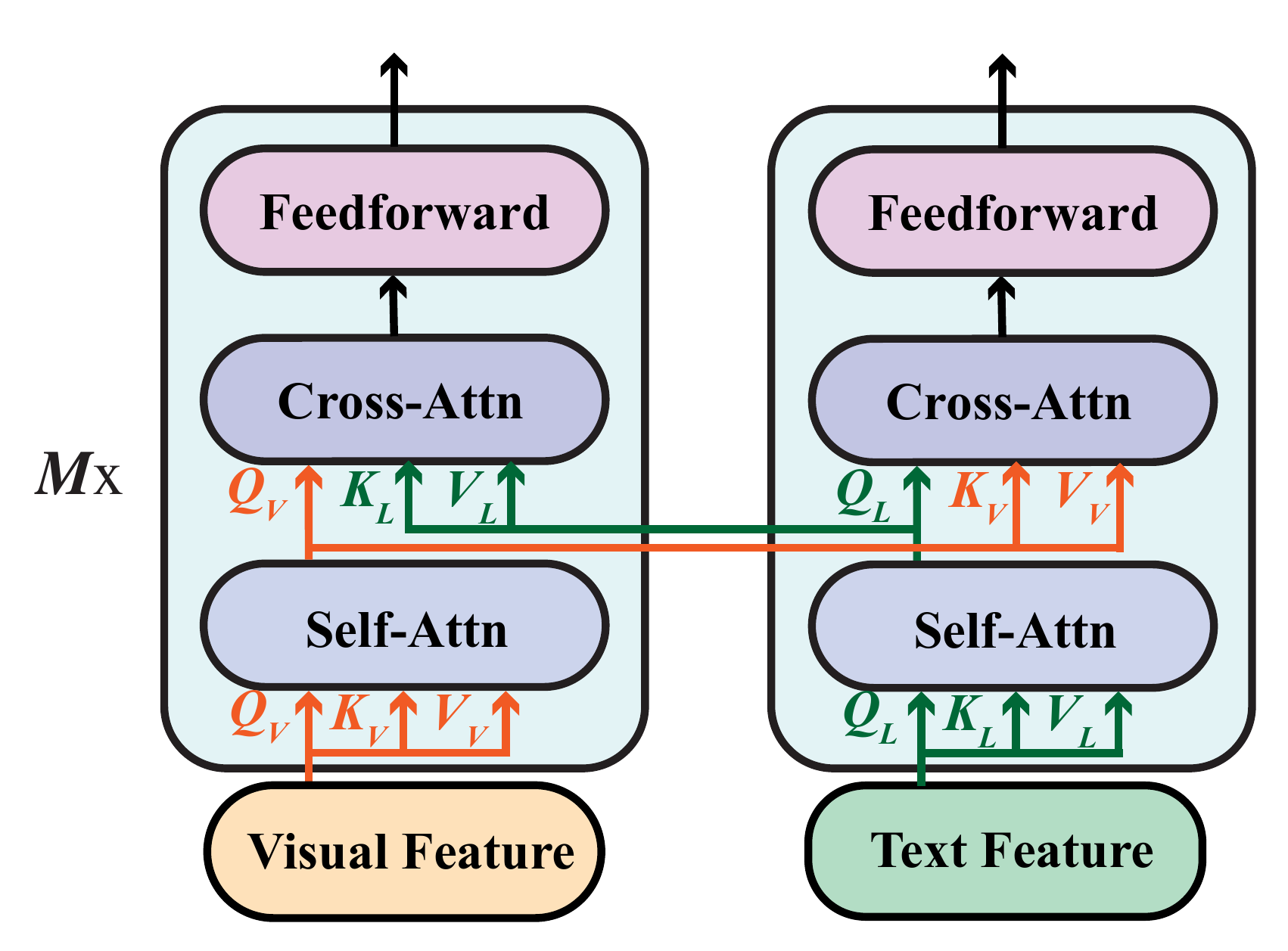}
    \caption{Co-attention model.}
    \label{fig:dual}
  \end{subfigure}
  \begin{subfigure}{0.49\linewidth}
    \includegraphics[width=1.0\linewidth]{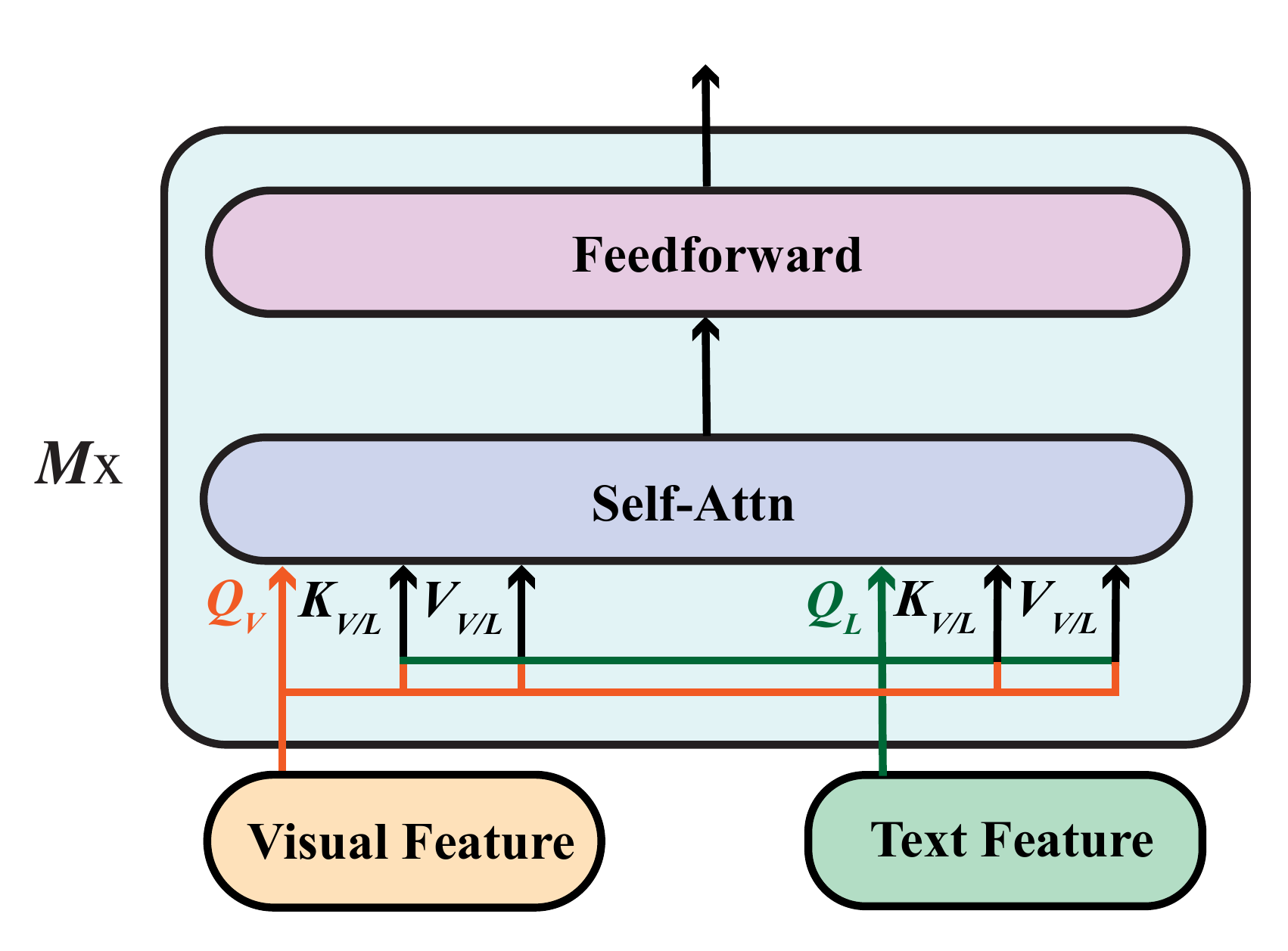}
    \caption{Merged attention model.}
    \label{fig:single}
  \end{subfigure}
  \caption{Illustration of two types of multimodal fusion modules: (a) co-attention, and (b) merged attention.}
  \label{fig:fusion_modules}
  \vspace{-6mm}
\end{figure}

\paragraph{Multimodal Fusion.}
We study two types of fusion modules, namely, \textit{merged attention} and \textit{co-attention}~\cite{hendricks2021decoupling}, as illustrated in Figure~\ref{fig:fusion_modules}.
In the \textit{merged attention} module, the text and visual features are simply concatenated together, then fed into a single transformer block. In the \textit{co-attention} module, on the other hand, the text and visual features are fed into different transformer blocks independently, and techniques such as cross-attention are used to enable cross-modal interaction. For region-based VLP models, as shown in \cite{bugliarello-etal-2020-multimodal}, the \textit{merged attention} and \textit{co-attention} models can achieve comparable performance.
Yet, the \textit{merged attention} module is more parameter-efficient, as the same set of parameters are used for both modalities.
Since end-to-end VLP models are becoming increasingly popular, in this work, we re-examine the impact of both types of fusion modules in our new context. 

\begin{figure}[t!]
  \centering
  \begin{subfigure}{0.34\linewidth}
    \includegraphics[width=1.0\linewidth]{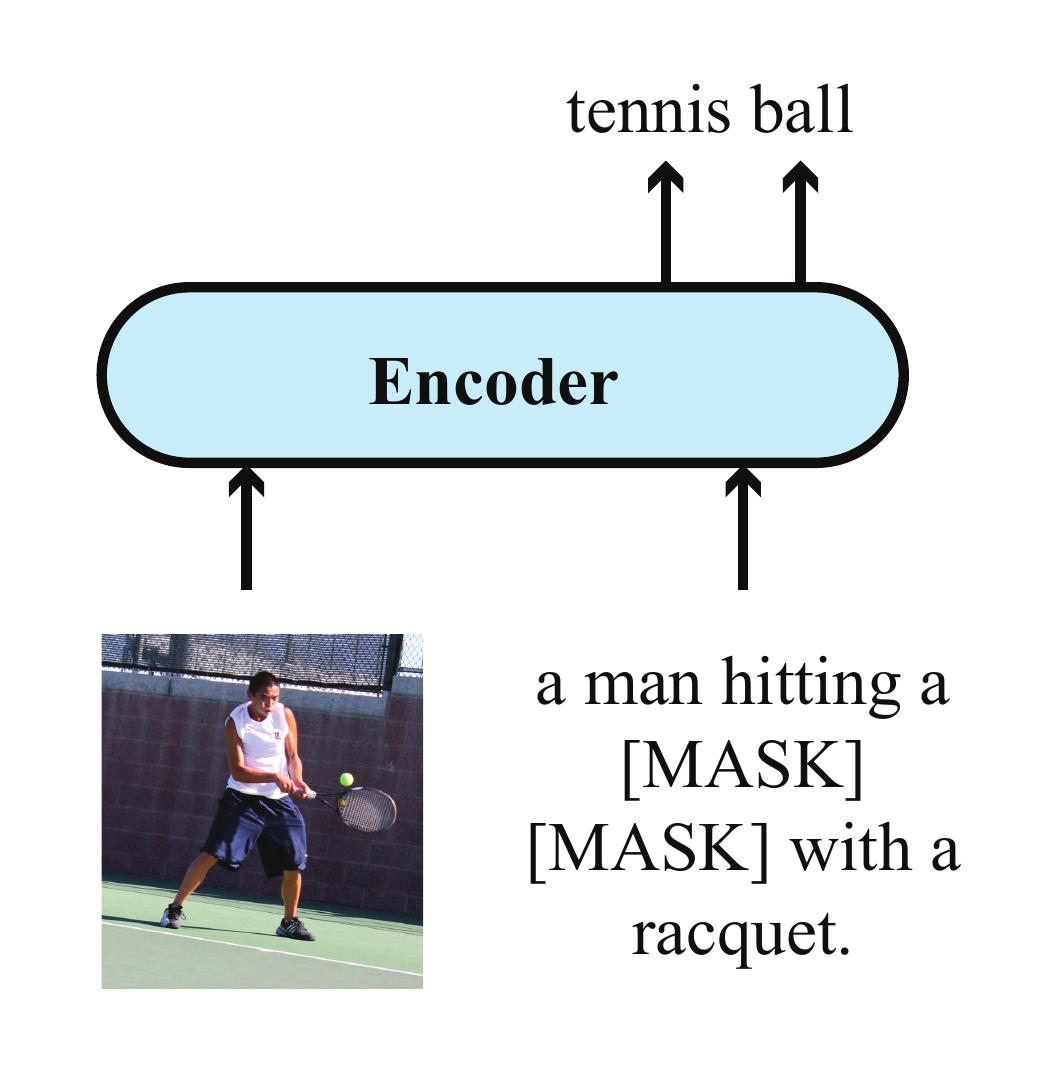}
    \caption{Encoder-Only}
    \label{fig:enc-only}
  \end{subfigure}
  \begin{subfigure}{0.59\linewidth}
    \includegraphics[width=1.0\linewidth]{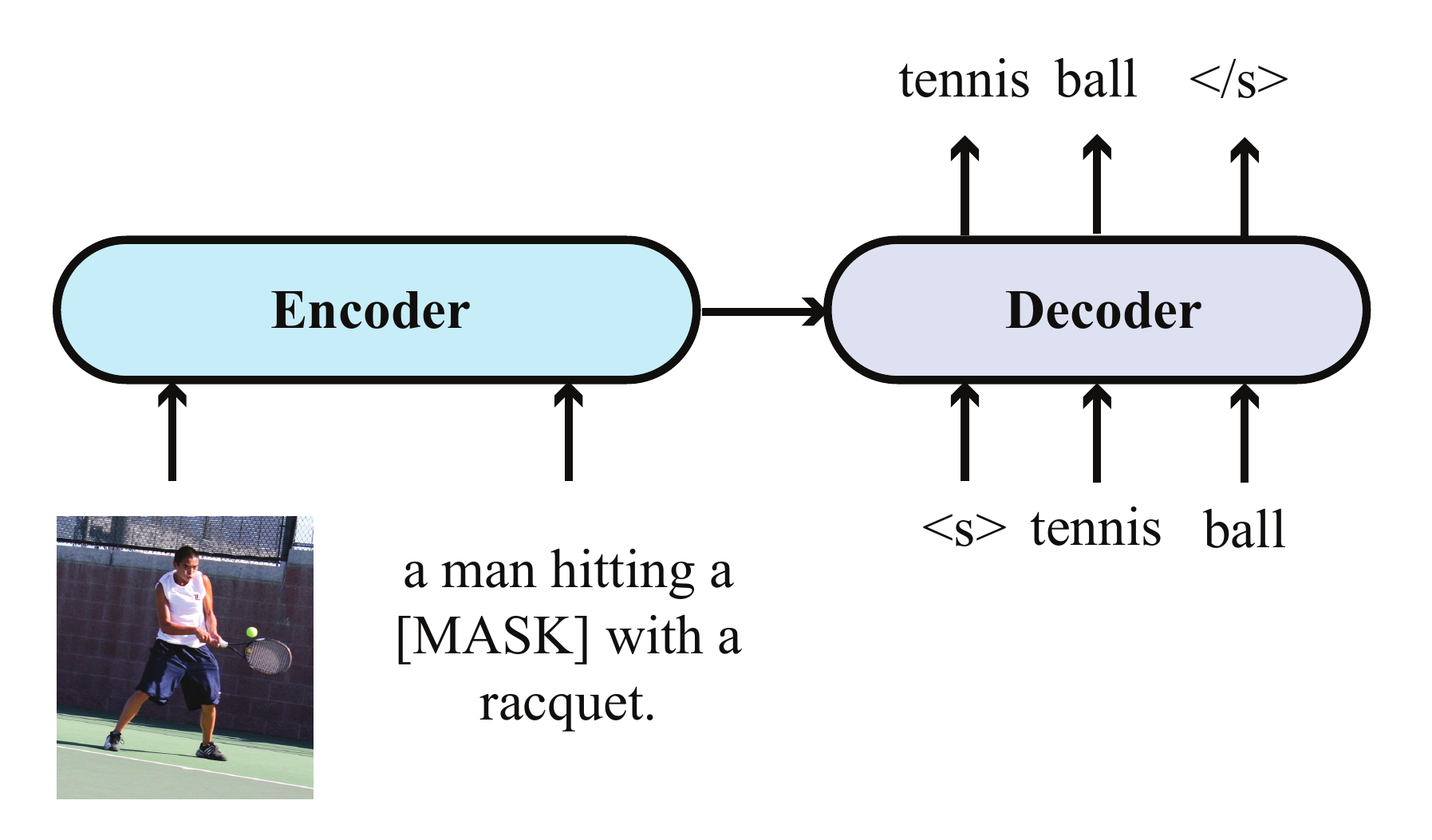}
    \caption{Encoder-Decoder}
    \label{fig:enc-dec}
  \end{subfigure}
  \caption{Illustration of the \textit{encoder-only} and \textit{encoder-decoder} model architectures for VLP.}
  \label{fig:enc_vs_enc_dec}
  \vspace{-6mm}
\end{figure}

\paragraph{Encoder-Only vs. Encoder-Decoder.} Many VLP models such as VisualBERT~\cite{li2019visualbert} adopt the encoder-only architecture, where the cross-modal representations are directly fed into an output layer to generate the final outputs. Recently, VL-T5~\cite{cho2021unifying} and SimVLM~\cite{wang2021simvlm}, on the other hand, advocate the use of a transformer encoder-decoder architecture, where the cross-modal representations are first fed into a decoder and then to an output layer. In their models, the decoder attends to both the encoder representations and the previously generated tokens, producing the outputs autoregressively. Figure~\ref{fig:enc_vs_enc_dec} shows the difference between them when performing the masked language modeling task. 
For the encoder-decoder model, when performing classification tasks such as VQA, we feed the text inputs into its encoder and feed a classification token into the decoder, and the decoder then generates the output class accordingly.

\subsection{Pre-training Objectives}
Now, we introduce how we pre-train our models. Specifically, we will first briefly review masked language modeling and image-text matching, which have been used extensively in almost every VLP model. Then, we will shift our focus to how we can design and explore interesting masked image modeling tasks. 


\paragraph{Masked Language Modeling.} The masked language modeling (MLM) objective is first introduced in pure language pre-training~\cite{devlin2018bert,liu2019roberta}. In VLP, MLM with images has also proven to be useful. Specifically, given an image-caption pair, we randomly mask some of the input tokens, and the model is trained to reconstruct the original tokens given the masked tokens $\mathbf{l}^{mask}$ and its corresponding visual input $\mathbf{v}$.

\paragraph{Image-Text Matching.} In image-text matching, the model is given a batch of matched or mismatched image-caption pairs, and the model needs to identify which images and captions correspond to each other. Most VLP models treat image-text matching as a binary classification problem. Specifically, a special token (\eg, $\texttt{[CLS]}$) is
inserted at the beginning of the input sentence, and it tries to learn a global cross-modal representation. We then feed the model with either a matched or mismatched image-caption pair $\langle \mathbf{v}, \mathbf{l} \rangle$ with equal probability, and a classifier is added on top of the $\texttt{[CLS]}$ token to predict
a binary label $y$, indicating if the sampled image-caption pair is a match. 

\paragraph{Masked Image Modeling.} Similar to the MLM objective, researchers have tried masked image modeling (MIM) on the vision side. For example, many previous work, such as LXMERT~\cite{tan-bansal-2019-lxmert} and UNITER~\cite{chen2020uniter}, mask some of the input regions, and the model is trained to regress the original region features. 
Formally, given a sequence of visual features $\mathbf{v}=\langle v_1, \cdots, v_M\rangle$, where $v_i$ is typically a region feature, we randomly mask some of the visual features, and the model outputs the reconstructed visual features $\mathbf{o_v}$ given the rest of the visual features and the unmasked tokens $\mathbf{t}$, and regression aims to minimize the mean squared error loss. 
Researchers~\cite{tan-bansal-2019-lxmert,lu2019vilbert,chen2020uniter} have also tried to first generate object label for each region using a pre-trained object detector, which can contain high-level semantic information, and the model is trained to predict the object labels for the masked regions instead of the original region features.

Notably, recent state-of-the-art models (\emph{e.g.}, ALBEF~\cite{li2021align}, VinVL~\cite{zhang2021vinvl}) do not apply MIM during VLP. In addition, in ViLT~\cite{kim2021vilt}, the authors also demonstrate that masked patch regression is not helpful in their setting. These results make it questionable whether MIM is truly effective for VLP models.
To further investigate this, we treat masked image modeling as a masked patch classification task, and propose two ways of implementing the idea.

\begin{figure}[t]
  \centering
   \includegraphics[width=0.98\linewidth]{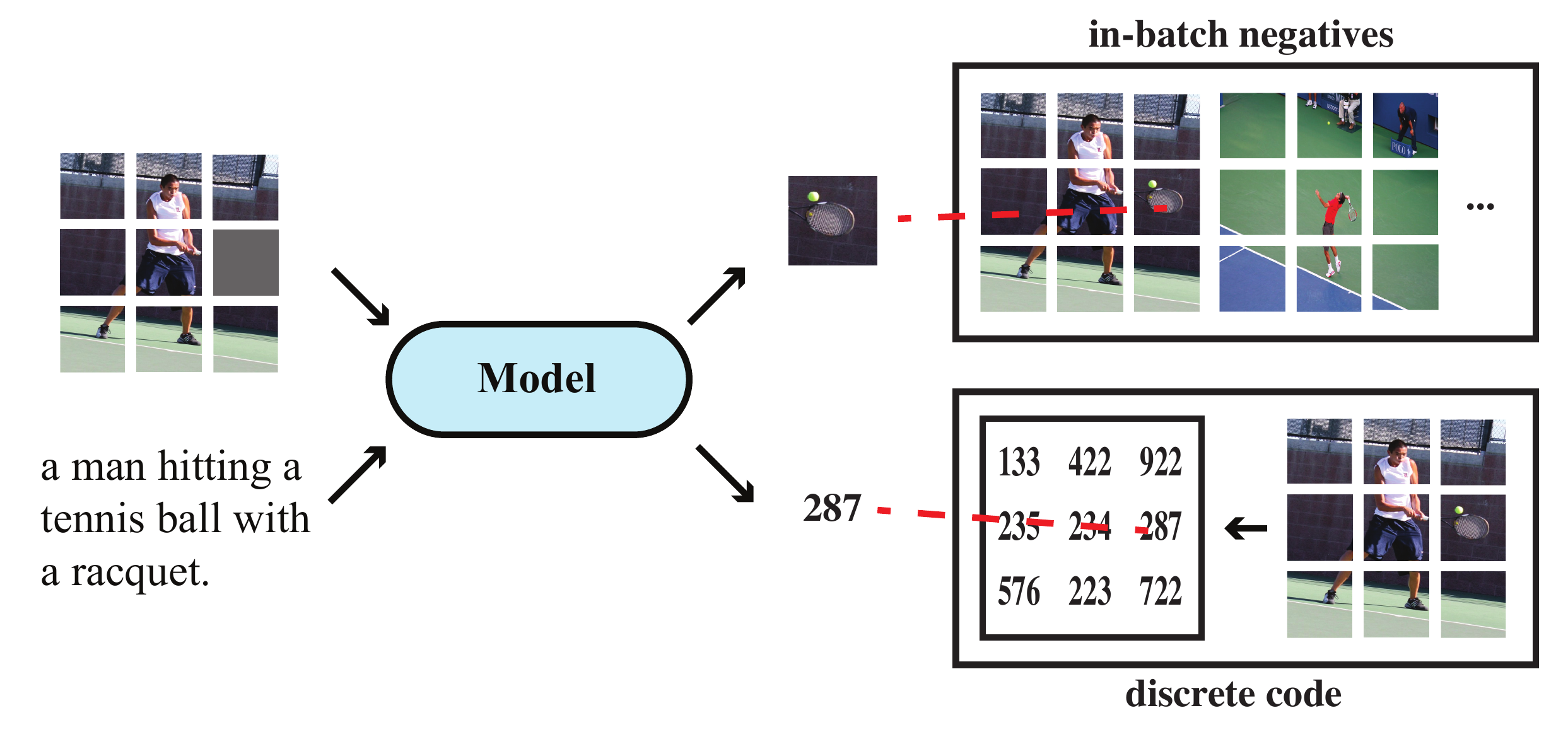}
   \vspace{-2mm}
   \caption{Illustration of masked patch classification with \textit{in-batch negatives} and with \textit{discrete code}.}
   \label{fig:mim}
   \vspace{-4mm}
\end{figure}
\paragraph{1) Masked Patch Classification with In-batch Negatives.}
By imitating MLM which uses a text vocabulary, we first propose to let the model reconstruct input patches by using a dynamically constructed vocabulary constructed with in-batch negatives. 
Concretely, at each training step, we sample a batch of image-caption pairs $\{ \langle \mathbf{v}^k, \mathbf{l}^k \rangle \}_{k=1}^B$, where $B$ is the batch size. We treat all the patches in $\{ \mathbf{v}^k \}_{k=1}^B$ as candidate patches, and for each masked patch, we mask 15\% of the input patches, and the model needs to select the original patch within this candidate set. Denoting the original patch representations and our model's output representations of $\{ \mathbf{v}^k \}_{k=1}^B$ as $\{ c(\mathbf{v}^k) \}_{k=1}^B$ and $\{ h(\mathbf{v}^k) \}_{k=1}^B$, respectively, we can represent the output probability at position $i$ for the $k$-th instance as: 
\begin{equation}
   p(\mathbf{v}_i^k|[\mathbf{v}^{k,mask}; \mathbf{l}^{k}]) = \frac{e^{h(\mathbf{v}_i^k)^{\mathrm{T}}c(\mathbf{v}_i^k)}}{\sum_{j, k'} e^{h(\mathbf{v}_i^k)^{\mathrm{T}}c(\mathbf{v}_j^{k'})}}. 
\end{equation}
The model is trained to maximize its probability similar to noise contrastive estimation~\cite{nce1,nce2}.

\paragraph{2) Masked Patch Classification with Discrete Code.} Inspired by BEiT~\cite{bao2021beit}, we also propose to obtain discrete representations of the input patches, and the model is then trained to reconstruct the discrete tokens.
Specifically, we first use the VQ-VAE~\cite{van2017neural} model in DALL-E~\cite{ramesh2021zero} to tokenize each image into a sequence of discrete tokens. We resize each image so that the number of patches is equal to the number of tokens, and thus each patch corresponds to a discrete token. Then, we randomly mask 15\% of the patches and feed the masked image patches to the model as before, but now the model is trained to predict the discrete tokens instead of the masked patches.

\subsection{Our Default Settings for \textbf{\ModelName}}\label{sec: default_settings}
There are many different model designs under \ModelName, and we detail our default settings in this part.

\paragraph{Model Architecture.} The default setting of model architecture is shown in~Figure \ref{fig:dual}. In the bottom part, there are one pre-trained visual encoder and one pre-trained text encoder. On top of each encoder, we stack $M=6$ transformer encoding layers, with each layer consisting of one self-attention block, one cross-attention block, and one feed-forward network block. 
Unless otherwise stated, the hidden size is set to $768$, and the number of heads is set to $12$ for the top layers. Note that there is no decoder and no parameter sharing between the vision and language branches.

\paragraph{Pre-training Objectives.} Unless otherwise stated, we pre-train the models with masked language modeling (MLM) and image-text matching (ITM) only. For MLM, we mask 15\% of the input text tokens, and the model tries to reconstruct the original tokens. For ITM, we feed the model with either matched or mismatched image-caption pairs with equal probability, and the model needs to identify whether the input is a match.

\paragraph{Pre-training Datasets.} Following previous work~\cite{chen2020uniter,kim2021vilt,li2021align}, we pre-train models on four commonly used datasets, including COCO~\cite{lin2014microsoft}, Conceptual Captions~\cite{sharma2018conceptual}, SBU Captions~\cite{ordonez2011im2text}, and Visual Genome~\cite{krishna2016visual}. The statistics of these datasets is shown in Appendix. The combined training data consists of about 4M images in total. 

\paragraph{Downstream Tasks.}
For ablation and analysis, we mainly focus on VQAv2~\cite{antol2015vqa}, arguably the most popular dataset for VLP evaluation. We also test on Flickr30k zero-shot image-text retrieval to remove any confounders that may be introduced during finetuning~\cite{hendricks2021decoupling}. For VQAv2, we follow the standard practice~\cite{chen2020uniter} to train the models with both training and validation data, and test the models on the test-dev set.
For Flickr30k, we follow the standard splits.

For comparison with state of the arts, we also evaluate models on visual reasoning (NLVR$^2$~\cite{suhr2018corpus}), visual entailment (SNLI-VE~\cite{xie2019visual}), and image-text retrieval (COCO~\cite{lin2014microsoft} and Flickr30k~\cite{plummer2015flickr30k}) tasks. For retrieval tasks, we evaluate models in both zero-shot and finetuning settings.

\paragraph{Implementation Details.} We pre-train our models using AdamW~\cite{loshchilov2018decoupled} for 100k steps. The learning rates for the bottom and top layers are set to 1e-5 and 5e-5 respectively during pre-training. The warm-up ratio is set to 10\%, and the learning rate is linearly decayed to 0 after 10\% of the total training steps. We use center-crop to resize each image into the size of 224$\times$224 or 384$\times$384 depending on the adopted vision transformers. 


 \vspace{-2mm}
\section{Experiments}
 \vspace{-1mm}
In this section, we provide comprehensive analysis of each individual module design. Specifically, ($i$) we study the impact of vision and language encoders in Section~\ref{sec:vision_lan_encoders}, ($ii$) we perform analysis on multimodal fusion designs in Section~\ref{sec:multimodal_fusion}, ($iii$) we compare encoder-only and encoder-decoder architectures in Section~\ref{sec:enc_vs_enc_dec}, and ($iv$) we ablate pre-training objectives in Section~\ref{sec:pretrain_objectives}. Finally, we compare with state of the arts in Section~\ref{sec: compare-sota}.

\begin{table}[!t]
\tablestyle{5pt}{1.0}
\def\w{20pt} 
\scalebox{1.0}{
  \begin{tabular}{l|cccc|cc}
    \bf \multirow{2}{*}{Text Enc.} & \bf VQAv2 & \bf VE & \bf IR & \bf TR  & \bf SQuAD & \bf MNLI\\
    & Acc. & Acc. & R@1 & R@1  & EM & Acc. \\
\shline
Emb-only & 67.13 & 74.85 & 49.06 & 68.20 & - & -\\
ELECTRA & 69.22 & 76.57 & 41.80 & 58.30 & 86.8 & \bf 88.8\\
CLIP & 69.31 & 75.37 & \bf 54.96 & \bf 73.80 & - & -\\
DeBERTa & 69.40 & \bf 76.74 & 51.50 & 67.70 & \bf 87.2 &  \bf 88.8\\
    BERT & 69.56 & 76.27 &49.60 & 66.60 & 76.3 &  84.3 \\
RoBERTa & 69.69 & 76.53 & 49.86 & 68.90& 84.6 & 	87.6\\
ALBERT & \bf 69.94 & 76.20& 52.20 & 68.70 & 86.4 & 87.9\\
 \end{tabular}
  }
  \caption{\textbf{Comparisons of different text encoders without VLP}. CLIP-ViT-224/32 is used as the vision encoder. All the text encoders are in base model size, except ALBERT, which is xlarge. Emb-only: only using word embeddings as text encoder. IR/TR: Flickr30k image/text retrieval. EM: exact match. The results of SQuAD and MNLI are copied from their corresponding papers. All the results on VL tasks are from their test-dev/val sets. }
  \label{tab:ablate_text_encoder}
  \vspace{-3mm}
\end{table}
\begin{table}[!t]
\tablestyle{5pt}{1.0}
\def\w{18pt} 
\scalebox{1.0}{
  \begin{tabular}{c|cccc|c}
    \bf Vision Encoder & \bf VQAv2  & \bf VE & \bf IR & \bf TR & \bf ImageNet\\
    \shline
     Dis. DeiT B-384/16 & 67.84 & 76.17 & 34.84 & 52.10 & 85.2\\
       BEiT B-224/16 & 68.45 & 75.28 & 32.24 & 59.80& 85.2\\
        DeiT B-384/16 & 68.92 & 75.97 & 33.38 & 50.90& 82.9\\
    ViT B-384/16 & 69.09 & 76.35 & 40.30 & 59.80 & 83.97 \\
     CLIP B-224/32 & 69.69 & 76.53 & 49.86 & 68.90 & -\\
     VOLO 4-448/32 & 71.44 & 76.42 & 40.90 & 61.40 & \bf 86.8\\
 CaiT M-384/32 & 71.52 & 76.62& 38.96 & 61.30 & 86.1\\
  CLIP B-224/16 & 71.75 & 77.54 & \bf 57.64 & \bf 76.90& -\\
 Swin B-384/32 & \bf 72.38 &  \bf 77.65 & 52.30 & 69.50 & 86.4 \\
  \end{tabular}
  }
  \caption{\textbf{Comparisons of different vision encoders without VLP}. RoBERTa is used as the default text encoder. IR/TR: Flickr30k image/text retrieval; B: Base. The results of ImageNet classification are copied from their corresponding papers. All the results on VL tasks are from their test-dev/val sets. N and M in ViT-N/M denote the image resolution and patch size, respectively.}
  \label{tab:abalte_vision_encoder}
  \vspace{-6mm}
\end{table}

\subsection{Impact of Vision and Language Encoders} \label{sec:vision_lan_encoders}


\subsubsection{Explorations without VLP} 
Since pre-training is time-consuming, we first perform an exploration study by comparing different text and visual encoders without VLP for efficiency. Concretely, we initialize the bottom layers with specific pre-trained vision and text encoders, and randomly initialize the top layers. Then, we directly finetune the models on three tasks, including VQAv2, SNLI-VE, and Flickr30k retrieval. The learning rates for the bottom and top layers are set to 1e-5 and 1e-4, and the number of training epochs is set to 10 for all the tasks.
We choose CLIP-ViT-224/32~\cite{radford2021learning} and RoBERTa~\cite{liu2019roberta} as the default encoders. Here, N and M in ViT-N/M denote image resolution and patch size, respectively.

\paragraph{Impact of Text Encoders.}
As shown in Table~\ref{tab:ablate_text_encoder}, there are no significant differences between the model performance of different text encoders. RoBERTa seems to achieve the most robust performance in this setting. Also, as can be seen from the Emb-only results, it is necessary to have a pre-trained encoder because otherwise the downstream task performance will be degraded.

\paragraph{Impact of Vision Encoders.} As summarized in~Table \ref{tab:abalte_vision_encoder}, both CLIP-ViT-224/16 and Swin Transformer can achieve decent performance in this setting. Notably, Swin Transformer can achieve an VQA score of \textbf{72.38} on the test-dev set \emph{\textbf{without any VLP}}, which is already comparable to some VLP models after pre-training.

\paragraph{Conclusion.}
If we directly finetune the models on downstream tasks without any VLP, RoBERTa and Swin Transformer or CLIP-ViT perform the best. While models such as DeBERTa and BEiT can achieve better performance than the two models on pure language or vision tasks such as MNLI~\cite{wang2018glue} or ImageNet classification~\cite{deng2009imagenet}, that does not necessarily indicate that they are more suitable for VL tasks.

\begin{table}[!t]
\tablestyle{5pt}{1.0}
\def\w{20pt} 
\scalebox{1.0}{
  \begin{tabular}{cc|ccc}
    \multirow{2}{*}{\bf Text Enc.} & \multirow{2}{*}{\bf Vision Enc.} &  \multirow{2}{*}{\bf VQAv2} &  \multicolumn{2}{c}{\bf Flickr-ZS} \\
     & & & \bf IR & \bf TR \\
    \shline
    Emb-only & CLIP-32 & 73.99 & 60.32 & 74.10 \\
    \hline
    \multirow{2}{*}{BERT}  &  CLIP-32 & 74.98 & 66.08 & 78.10\\
    & CLIP-16 &  76.70 &  74.52& 87.20 \\
    \hline
    \multirow{3}{*}{RoBERTa}  &  CLIP-32 & 74.67 & 65.50  & 76.60  \\
    & CLIP-16 & \bf 77.19 & \bf 76.64 & \bf 89.60 \\
    & Swin &  76.43 & 71.68 & 85.30 \\
  \end{tabular}
  }
  \caption{\textbf{Comparisons of different vision and text encoders with VLP}. Results on VQAv2 are on test-dev set. ZS: zero-shot.}
  \label{tab:ablation_encoders_with_vlp}
   \vspace{-6mm}
\end{table}

\subsubsection{Results with VLP}
\label{vlp:1}

Now, we follow the default setting in Section~\ref{sec: default_settings}, and compare different vision/text encoders with VLP. Based on the previous results, we compare Embed-only, BERT, and RoBERTa on the text side, and CLIP-ViT-224/32, CLIP-ViT-224/16, and Swin Transformer on the image side. 


\paragraph{Results.} As shown in Table \ref{tab:ablation_encoders_with_vlp}, after VLP, the difference between BERT and RoBERTa seems to be diminished, but it is still important to have a pre-trained text encoder on the bottom (Embed-only vs. RoBERTa). For vision encoder, both CLIP-ViT-224/16 and Swin Transformer can achieve pretty good performance. Especially, CLIP-ViT-224/16 can achieve a VQA score of 77.19/77.20 on the test-dev/test-std sets, respectively, outperforming the previous state-of-the-art region-based VinVL~\cite{zhang2021vinvl} models. 

\begin{table}[!t]
\tablestyle{5pt}{1.0}
\def\w{20pt} 
\scalebox{1.0}{
  \begin{tabular}{cc|ccc}
    \multirow{2}{*}{\bf Bottom LR} & \multirow{2}{*}{\bf Top LR} &  \multirow{2}{*}{\bf VQAv2} &  \multicolumn{2}{c}{\bf Flickr-ZS} \\
     & & & \bf IR & \bf TR \\
    \shline
    1e-5 & 1e-5 & 73.16 & 48.80 & 63.70\\
    2e-5 & 2e-5 & 73.66 & 53.14 & 67.20 \\
    3e-5 & 3e-5 & 73.77 & 56.48 & 70.90\\
    5e-5 & 5e-5 & 73.54 & 52.48 & 65.90\\
    1e-5 & 5e-5 & \bf 74.98 & \bf 66.08 & \bf 78.10 \\
  \end{tabular}
  }
  \caption{Using different learning rates for the randomly-initialized and pre-trained parameters is better than using the same learning rate. Results on VQAv2 are on test-dev set. ZS: zero-shot.}
  \label{tab:lr_trick}
  \vspace{-3mm}
\end{table}
\paragraph{Useful Tricks.} In experiments, we found two tricks for ViT-based VLP models that can greatly boost the performance.
First, it is better to use a \emph{larger} learning rate for the randomly initialized parameters than parameters initialized with pre-trained models, which is also found useful in some other NLP tasks~\cite{liu2019text}. As shown in Table \ref{tab:lr_trick}, using the same learning rate for all parts of the model will lead to degraded performance, possibly because the pre-trained parameters already contain certain amounts of knowledge about vision and language, and finetuning them aggressively can result in the loss of these valuable information. 

\begin{figure}[t!]
  \centering
   \includegraphics[width=0.7\linewidth]{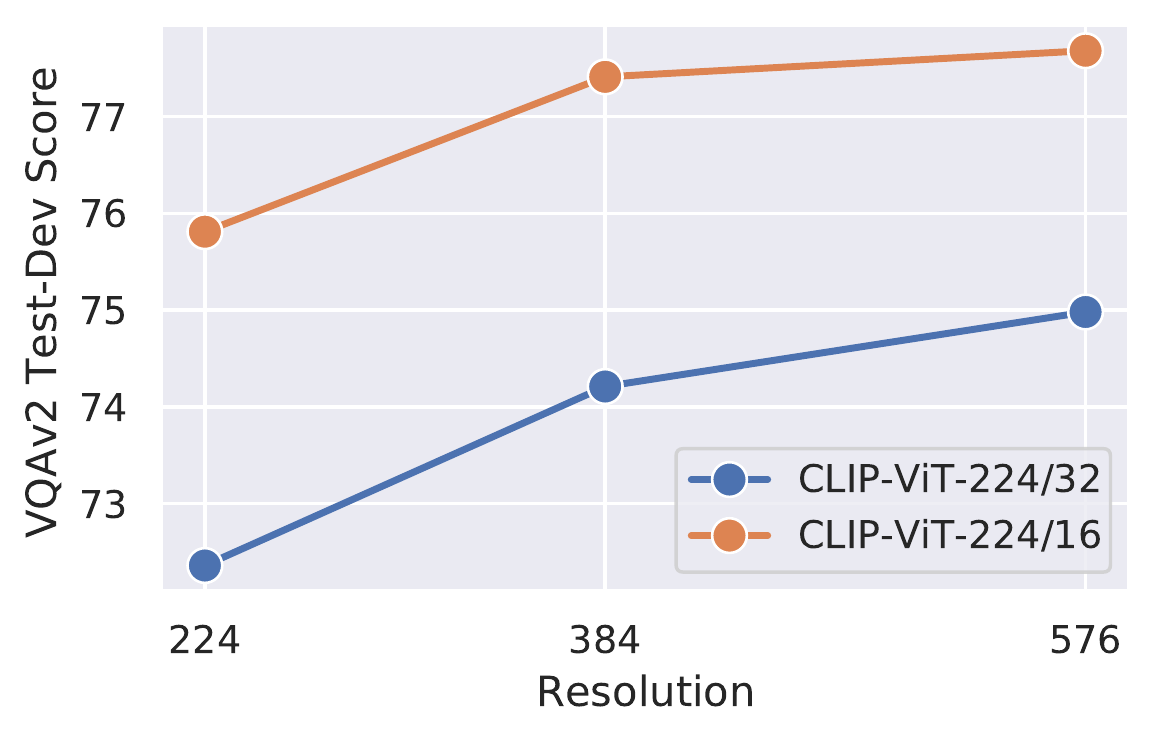}
   \vspace{-3mm}
   \caption{Increasing the image resolution during finetuning greatly improves the performance on the VQAv2 test-dev set. 
   }
   \label{fig:resolution}
   \vspace{-3mm}
\end{figure}

Second, similar to several previous work~\cite{yuan2021volo,kim2021vilt}, we find that increasing the image resolution during finetuning can improve the model performance by a large margin, especially when the ratio of image resolution to patch size is low. Figure~\ref{fig:resolution} shows that increasing the image resolution from 224 to 576 can improve the VQA score by about 3 and 1 points for the CLIP-ViT-224/32 and CLIP-ViT-224/16 model, respectively. 





\subsection{Analysis of the Multimodal Fusion Module}\label{sec:multimodal_fusion}
Now, following the default setting in Section~\ref{sec: default_settings},
we perform investigations on multimodal fusion.
First, we design both \emph{merged attention} and \emph{co-attention} models and investigate their performance.
For the merged attention model (Figure~\ref{fig:single}), 
the top transformer consists of $M_{merged}$ encoding layer, with each layer consisting of one self-attention block and one feed-forward network block. To help the model distinguish between the two modalities, we add a modality embedding to the input features before feeding them to the top transformer.
For the co-attention model (Figure~\ref{fig:dual}), we feed the text and visual features to two $M_{co}$-layer transformers separately, and each top transformer encoding layer consists of one self-attention block, one cross-attention block, and one feed-forward network block. Compared with merged attention, co-attention allows separate transformation functions for the vision and language modalities.
We set $M_{merged}=12$ and $M_{co}=6$ so that the numbers of parameters of the two models are roughly comparable to each other. 
\begin{table}[!t]
\tablestyle{5pt}{1.0}
\def\w{20pt} 
\scalebox{1.0}{
  \begin{tabular}{lc|ccc}
    \multirow{2}{*}{\bf Fusion  } & \multirow{2}{*}{\bf Decoder}   &  \multirow{2}{*}{\bf VQAv2} &  \multicolumn{2}{c}{\bf Flickr-ZS} \\
     &  & & \bf IR & \bf TR \\
    \shline
    Merged attention & \multirow{2}{*}{ \ding{55}} &74.00 & 57.46 & 73.10\\
    \cdashline{1-1} 
   \multirow{2}{*}{Co-attention} &  & \bf 74.98 & \bf 66.08 & \bf 78.10 \\
    \cdashline{2-2}
      & \ding{51} & 74.73 & 48.96 & 71.60 \\
  \end{tabular}
  }
  \caption{Co-attention performs better than merged attention in our setting, and adding a decoder is not helpful for our discriminative VL tasks. Results on VQAv2 are on test-dev set. ZS: zero-shot. }
  \label{tab:ablate_multimodal_fusion}
  \vspace{-6mm}
\end{table}

\paragraph{Results.} Table~\ref{tab:ablate_multimodal_fusion} reports the downstream performance of the two models. The co-attention model performs better than the merged attention model in our setting, indicating that it is important to have different sets of parameters for the two modalities. Note that this contradicts with the findings in region-based VLP models~\cite{bugliarello-etal-2020-multimodal}, possibly because ($i$) findings of region-based VLP models cannot directly apply to ViT-based VLP models; ($ii$) most region-based VLP models only use pre-trained visual encoders, and also do not have a pre-trained text encoder included, thus the inconsistency between the two modalities will not favor symmetrical architecture like the co-attention model. 

\subsection{Encoder-Only vs. Encoder-Decoder}
\label{sec:enc_vs_enc_dec}

We then compare the encoder-only and encoder-decoder architecture. For the encoder-only model, we use the same co-attention model as in Section \ref{sec:multimodal_fusion}. For the encoder-decoder model, we set the number of layers to 3 for both the encoder and decoder, and each decoding layer has two separate cross-attention blocks that attend to the vision and text representations, respectively. According to \cite{cho2021unifying}, we adopt T5-style~\cite{raffel2020exploring} language modeling objective as it works well for their model. Specifically, we mask 15\% of
input text tokens and replace contiguous text span with
sentinel tokens, and the decoder is trained to reconstruct the masked tokens. For image-text matching, we feed the decoder with a special class token and it generates a binary output.

\begin{table}[!t]
\tablestyle{5pt}{1.0}
\def\w{20pt} 
\scalebox{1.0}{
  \begin{tabular}{l|ccc}
    \multirow{2}{*}{\bf Pre-training Objectives}  &  \multirow{2}{*}{\bf VQAv2} &  \multicolumn{2}{c}{\bf Flickr-ZS} \\
     & & \bf IR & \bf TR \\
    \shline
    MLM & 74.19 & - & - \\
    ITM & 72.63  & 53.74 & 71.00\\
    MLM+ITM & \bf 74.98 & \bf 66.08 & \bf 78.10 \\
    MLM+ITM + MIM (In-batch Negatives) & 74.01 & 62.12 & 76.90\\
    MLM+ITM + MIM (Discrete Code) & 74.21 & 59.80 & 76.30\\
  \end{tabular}
  }
  \caption{Masked language modeling (MLM) and image-text matching (ITM) can both improve the model performance, but both of our designed masked image modeling (MIM) objectives lead to degraded performance on downstream tasks. Results on VQAv2 are on test-dev set. ZS: zero-shot.}
  \vspace{-3mm}
  \label{tab:mim}
\end{table}

\begin{table*}[!t]
\tablestyle{5pt}{1.0}
\def\w{20pt} 
\scalebox{1.0}{
  \begin{tabular}{l|cccccc|cccccc}
    \multirow{2}{*}{\bf Model}  &  \multicolumn{2}{c}{\bf VQAv2} & \multicolumn{2}{c}{\bf NLVR$^2$} &  \multicolumn{2}{c}{\bf SNLI-VE} &  \multicolumn{6}{c}{\bf Flickr-ZS}  \\
     & \bf test-dev & \bf test-std & \bf dev & \bf test & \bf dev & \bf test & \bf IR@1  & \bf IR@5  & \bf IR@10 & \bf TR@1 & \bf TR@5 & \bf TR@10 \\
    \shline
  \multicolumn{13}{l}{ \demph{ \it{Pre-trained with $>$10M images} } }\\
    \hline
    \demph{ALBEF (14M)~\cite{li2021align}} & \demph{75.84} & \demph{76.04} & \demph{82.55} & \demph{83.14} & \demph{80.80} & \demph{80.91} & \demph{82.8} & \demph{96.3} & \demph{98.1} & \demph{94.1} & \demph{99.5} & \demph{99.7} \\
    \demph{SimVLM$_{\text{BASE}}$ (1.8B)~\cite{wang2021simvlm}} & \demph{77.87} & \demph{78.14} & \demph{81.72} & \demph{81.77} & \demph{84.20} & \demph{84.15} & \demph{-} & \demph{-} & \demph{-} & \demph{-} & \demph{-} & \demph{-} \\
    \demph{SimVLM$_{\text{HUGE}}$ (1.8B)~\cite{wang2021simvlm}} & \demph{80.03} & \demph{80.34} & \demph{84.53} & \demph{85.15} & \demph{86.21} & \demph{86.32} & \demph{-} & \demph{-} & \demph{-} & \demph{-} & \demph{-} & \demph{-} \\
    \hline
    \multicolumn{13}{l}{ { \it{Pre-trained with $<$10M images} } }\\
    \hline
  UNITER$_{\text{LARGE}}$  ~\cite{chen2020uniter}  & 73.82 & 74.02  & 79.12 & 79.98 & 79.39 & 79.38 & 68.74 & 89.20 & 93.86 & 83.60 & 95.70 & 97.70 \\
  VILLA$_{\text{LARGE}}$~\cite{gan2020large}  & 74.69 & 74.87  & 79.76 & 81.47 & 80.18 & 80.02 & - & - & - & - & - & - \\
  UNIMO$_{\text{LARGE}}$~\cite{li2020unimo} & 75.06 & 75.27  & - & - & 
  \bf 81.11 & \underline{ 80.63} & - & - & - & - & - & - \\
  VinVL$_{\text{LARGE}}$~\cite{zhang2021vinvl} & \underline{ 76.52 }& 76.60 & \bf 82.67 & \bf 83.98 & - & - & -  & - & - & - & - & -  \\
  PixelBERT~\cite{huang2020pixel} & 74.45 & 74.55 & 76.5 & 77.2 & -  & - & - & - & - & - \\
  CLIP-ViL (ResNet50x4)~\cite{shen2021much} & 76.48 & \underline{ 76.70} & - & - & 80.61 & 80.20 & - & - & - & - & - & -\\
  \hline 
  ViLT~\cite{zhang2021vinvl} & 71.26 & - & 75.70 & 76.13 & - & - & 55.0 & 82.5 & 89.8 & 73.2 & 93.6 & 96.5 \\
  Visual Parsing~\cite{xue2021probing} & 74.00 & 74.17 & 77.61 & 78.05 & - & -  & - & - & - & - & - & -\\
  ALBEF (4M)~\cite{li2021align} & 74.54 & 74.70 & 80.24 & 80.50 & 80.14 & 80.30 & \underline{ 76.8} & \underline{ 93.7 }& \underline{96.7 }& \underline{90.5} & \bf 98.8 & \bf 99.7\\
  \ModelName-Swin$_{\text{BASE}}$  & 76.43 & 76.42 & 82.23 & 82.47 & 80.61 & 80.45 & 71.68 & 91.80 & 95.30 & 85.30 & 97.70 & 99.20\\
  \ModelName-CLIP-ViT$_{\text{BASE}}$  & \bf 77.68 & \bf 77.64 & \underline{82.33} & \underline{83.05} & \underline{ 80.86} & \bf 81.19 & \bf 79.60 & \bf 94.96 & \bf 97.28 & \bf 90.90 & \underline{98.30} &\underline{ 99.50} \\
  \end{tabular}
  }
  \caption{Comparisons with models pre-trained with $<$10M images on visual question answering, visual reasoning, visual entailment, and zero-shot image retrieval (IR) and text retrieval (TR) tasks. The best scores are in \textbf{bold}, and the second best scores are \underline{underlined}. }
  \label{tab:results1}
   \vspace{-3mm}
\end{table*}
\begin{table*}[!t]
\tablestyle{5pt}{1.0}
\def\w{20pt} 
\scalebox{1.0}{
  \begin{tabular}{l|cccccc|cccccc}
    \multirow{2}{*}{\bf Model}  & \multicolumn{6}{c}{\bf Flickr} & \multicolumn{6}{c}{\bf COCO}  \\
     &  \bf IR@1  & \bf IR@5  & \bf IR@10 & \bf TR@1 & \bf TR@5 & \bf TR@10 &  \bf IR@1  & \bf IR@5  & \bf IR@10 & \bf TR@1 & \bf TR@5 & \bf TR@10 \\
    \shline
    \multicolumn{13}{l}{ \demph{ \it{Pre-trained with $>$10M images} } }\\
    \hline
    \demph{ALBEF (14M)~\cite{li2021align}} & \demph{85.6} & \demph{97.5} & \demph{98.9} & \demph{95.9} & \demph{99.8} & \demph{100.0} & \demph{60.7} & \demph{84.3} & \demph{90.5} & \demph{77.6} & \demph{94.3} & \demph{97.2}\\
    \hline
    \multicolumn{13}{l}{ { \it{Pre-trained with $<$10M images} } }\\
    \hline
    UNITER$_{\text{LARGE}}$~\cite{chen2020uniter} & 75.56 & 94.08 & 96.76 & 87.30 & 98.00 & 99.20 & 52.93 & 79.93 & 87.95 & 65.68 & 88.56 & 93.76 \\
    VILLA$_{\text{LARGE}}$~\cite{gan2020large} & 76.26 & 94.24 & 96.84 & 87.90 & 97.50 & 98.80 & - & - & - & - & - & - \\
    UNIMO$_{\text{LARGE}}$~\cite{li2020unimo} & 78.04 & 94.24 & 97.12 & 89.40 & 98.90 & \underline{ 99.80} & - & - & - & - & - & - \\
  VinVL$_{\text{LARGE}}$~\cite{zhang2021vinvl} & - & - & - & - & - & -  & \bf 58.8 & \bf 83.5 & \bf 90.3 & \underline{ 75.4} & \underline{ 92.9} & \underline{ 96.2} \\
  PixelBERT~\cite{huang2020pixel} & 71.5 & 92.1 & 95.8 & 87.0 & 98.9 & 99.5 & 50.1 & 77.6 & 86.2 & 63.6 & 87.5 & 93.6 \\
  \hline 
  ViLT~\cite{zhang2021vinvl} & 64.4 & 88.7 & 93.8 & 83.5 & 96.7 & 98.6 & 42.7 & 72.9 & 83.1 & 61.5 & 86.3 & 92.7 \\
  Visual Parsing~\cite{xue2021probing} & 73.5 & 93.1 & 96.4 & 87.0 & 98.4 & 99.5 & - & - & - & - & - & - \\
  ALBEF (4M)~\cite{li2021align} & \bf 82.8 & \bf 96.7 & \bf 98.4 & \bf 94.3 & \underline{ 99.4} &\underline{ 99.8}  & 56.8 & 81.5 & 89.2 & 73.1 & 91.4 & 96.0\\
  \ModelName-Swin$_{\text{BASE}}$  & 79.02 & 95.58 & 98.04 & 92.40 & 99.00 & 99.50 & 54.85 & 81.41 & 89.31 & 72.96 & 92.02 & 96.26  \\
  \ModelName-CLIP-ViT$_{\text{BASE}}$  & \underline{82.22} &  \underline{96.34} & \bf 98.36 & \bf 94.30 & \bf 99.60 & \bf 99.90 & \underline{ 57.08} & \underline{ 82.66} & \underline{90.07} & \bf 76.16 & \bf 93.16 & \bf 96.82\\
  \end{tabular}
  }
  \caption{Comparisons with models pre-trained with $<$10M images on Flickr30k and COCO image retrieval (IR) and text retrieval (TR) tasks in the finetuning setting. The best scores are in \textbf{bold}, and the second best scores are \underline{underlined}. }
  \label{tab:results2}
   \vspace{-5mm}
\end{table*}

\paragraph{Results.} As shown in Table~\ref{tab:ablate_multimodal_fusion}, the encoder-only model can outperform the encoder-decoder model on our two discriminative tasks, which is consistent with the findings in \cite{cho2021unifying}. However, it should be noted that the encoder-decoder architecture is more flexible, as it can perform tasks such as image captioning which may not be that straightforward for an encoder-only model to be applied to.

\subsection{Ablations on Pre-training Objectives}
\label{sec:pretrain_objectives}
In all the previous experiments, we pre-train our models with different objectives, following the default setting in~Section \ref{sec: default_settings}. Now, we alter the pre-training objectives.

\paragraph{Results.} As summarized in~Table \ref{tab:mim}, both masked language modeling and image-text matching can bring performance improvements on downstream tasks. However, both of our masked image modeling objectives can lead to degraded performance on both VQAv2 and Flickr30k retrieval tasks. This further indicates that conclusions in region-based VLP models may not necessarily hold in vision transformer-based models. 
We hypothesize that the performance drop is due to the conflicts between different objectives, and some techniques in multi-task optimization~\cite{yu2020gradient,wang2020gradient} may be borrowed to resolve the conflicts, which we list as one of the future directions. Another possible reason is that image patches can be noisy, thus the supervisions on reconstructing these noisy patches can be uninformative.


\subsection{Comparison with Prior Arts}\label{sec: compare-sota}
In this section, we evaluate our best-performing models (\emph{i.e.}, RoBERTa-base+Swin Transformer and RoBERT-base+CLIP-ViT-224/16 with co-attention fusion module, and with image resolutions set to 384 and 288, respectively), and compare them with previous work. We evaluate the models on visual question answering (VQAv2), visual reasoning (NLVR$^2$), visual entailment (SNLI-VE), Flickr30k retrieval tasks in zero-shot and finetuning settings, and COCO retrieval tasks in the finetuning setting.

\paragraph{Main Results.} As in Table \ref{tab:results1} and \ref{tab:results2}, compared with models pre-trained with fewer than 10M images, our CLIP-based model (\ModelName-CLIP-ViT$_\text{BASE}$) can achieve either the best or the second best scores on all the downstream tasks. Notably, our model can achieve a VQA score of 77.64\% on the VQAv2 test-std set using only 4M images for pre-training, surpassing the state-of-the-art region-feature-based VinVL model by 1.04\%, and outperforming the previous best fully transformer-based model (\ie, ALBEF) by 1.6\%. In addition, while ALBEF has specially-designed objectives for retrieval, 
our model can still outperform ALBEF on text and image retrieval tasks, further demonstrating the effectiveness of \ModelName. Also, as shown in Appendix, we can maintain the fast inference speed of ViT-based models.

\begin{table}[!t]
\tablestyle{5pt}{1.0}
\def\w{25pt} 
\scalebox{1.0}{
  \begin{tabular}{l|cc}
   \bf Model (\#Pre-training Images) & \bf test-dev & \bf test-std \\ 
    \shline
      SimVLM$_{\text{BASE}}$ (1.8B) & 77.87 & 78.14 \\
      SimVLM$_{\text{HUGE}}$ (1.8B) & 80.03 & 80.34 \\
      \cdashline{1-3}
      \ModelName-CoSwin$_{\text{HUGE}}$  (14M) & \bf 80.33 & \bf 80.54\\
  \end{tabular}
  }
  \caption{Pre-training a huge model under the \ModelName framework with 14M images can lead to state-of-the-art performance on VQAv2, surpassing previous models trained with 1.8B images.}
  \label{tab:florence}
  \vspace{-5mm}
\end{table}

\paragraph{Scaling the Model.}
We also investigate if the \ModelName framework is scalable. To this end, we pre-train our model with more images and larger vision backbone. Specifically, we pre-train the model with COCO, CC, CC12M~\cite{changpinyo2021conceptual}, SBU, and VG datasets, consisting of about 14M images and 20M image-caption pairs in total. We use CoSwin-Huge~\cite{yuan2021florence} as our vision backbone and RoBERTa-base as our text backbone. The hidden size of the fusion module remains unchanged. As shown in Table~\ref{tab:florence}, our model can achieve state-of-the-art performance on VQAv2, surpassing previous models trained with 1.8B images. The results indicate that our \ModelName framework is scalable.

\paragraph{Further Analysis.} We also conduct experiments on image captioning, investigate multi-scale feature fusion,  study the model performance on unimodal tasks after VLP, and provide visualization of learned attention maps. All these results are provided in Appendix.

\section{Conclusion}
We present~\ModelName, through which we systematically investigate how to train a fully-transformer VLP model in an end-to-end manner. 
Experiments demonstrate that we can achieve competitive performance with state-of-the-art models with only 4M images for pre-training. When further scaled up, \ModelName achieves new state of the art on VQA. 

{\small
\bibliographystyle{ieee_fullname}
\bibliography{egbib}

\begin{thebibliography}{10}\itemsep=-1pt

\bibitem{antol2015vqa}
Stanislaw Antol, Aishwarya Agrawal, Jiasen Lu, Margaret Mitchell, Dhruv Batra,
  C Lawrence~Zitnick, and Devi Parikh.
\newblock {VQA}: Visual question answering.
\newblock In {\em International Conference on Computer Vision (ICCV)}, 2015.

\bibitem{bao2021beit}
Hangbo Bao, Li Dong, and Furu Wei.
\newblock {BEiT}: Bert pre-training of image transformers.
\newblock {\em arXiv preprint}, 2021.

\bibitem{bapna2018training}
Ankur Bapna, Mia~Xu Chen, Orhan Firat, Yuan Cao, and Yonghui Wu.
\newblock Training deeper neural machine translation models with transparent
  attention.
\newblock In {\em Conference on Empirical Methods in Natural Language
  Processing (EMNLP)}, 2018.

\bibitem{bugliarello-etal-2020-multimodal}
Emanuele Bugliarello, Ryan Cotterell, Naoaki Okazaki, and Desmond Elliott.
\newblock Multimodal pretraining unmasked: {U}nifying the vision and language
  {BERT}s.
\newblock {\em Transactions of the Association for Computational Linguistics
  (TACL)}, 2021.

\bibitem{changpinyo2021conceptual}
Soravit Changpinyo, Piyush Sharma, Nan Ding, and Radu Soricut.
\newblock Conceptual 12m: Pushing web-scale image-text pre-training to
  recognize long-tail visual concepts.
\newblock In {\em Conference on Computer Vision and Pattern Recognition
  (CVPR)}, 2021.

\bibitem{chen2020uniter}
Yen-Chun Chen, Linjie Li, Licheng Yu, Ahmed El~Kholy, Faisal Ahmed, Zhe Gan, Yu
  Cheng, and Jingjing Liu.
\newblock {UNITER}: Universal image-text representation learning.
\newblock In {\em European Conference on Computer Vision (ECCV)}, 2020.

\bibitem{cho2021unifying}
Jaemin Cho, Jie Lei, Hao Tan, and Mohit Bansal.
\newblock Unifying vision-and-language tasks via text generation.
\newblock In {\em International Conference on Machine Learning (ICML)}, 2021.

\bibitem{clark2020electra}
Kevin Clark, Minh-Thang Luong, Quoc~V Le, and Christopher~D Manning.
\newblock Electra: Pre-training text encoders as discriminators rather than
  generators.
\newblock In {\em International Conference on Learning Representations (ICLR)},
  2020.

\bibitem{cubuk2020randaugment}
Ekin~D Cubuk, Barret Zoph, Jonathon Shlens, and Quoc~V Le.
\newblock Randaugment: Practical automated data augmentation with a reduced
  search space.
\newblock In {\em Conference on Computer Vision and Pattern Recognition (CVPR)
  Workshops}, 2020.

\bibitem{deng2009imagenet}
Jia Deng, Wei Dong, Richard Socher, Li-Jia Li, Kai Li, and Li Fei-Fei.
\newblock Imagenet: A large-scale hierarchical image database.
\newblock In {\em Conference on Computer Vision and Pattern Recognition
  (CVPR)}, 2009.

\bibitem{devlin2018bert}
Jacob Devlin, Ming-Wei Chang, Kenton Lee, and Kristina Toutanova.
\newblock {BERT}: Pre-training of deep bidirectional transformers for language
  understanding.
\newblock In {\em Conference of the North {A}merican Chapter of the Association
  for Computational Linguistics (NAACL)}, 2019.

\bibitem{dosovitskiy2020image}
Alexey Dosovitskiy, Lucas Beyer, Alexander Kolesnikov, Dirk Weissenborn,
  Xiaohua Zhai, Thomas Unterthiner, Mostafa Dehghani, Matthias Minderer, Georg
  Heigold, Sylvain Gelly, et~al.
\newblock An image is worth 16x16 words: Transformers for image recognition at
  scale.
\newblock In {\em International Conference on Learning Representations (ICLR)},
  2021.

\bibitem{dou2018exploiting}
Zi-Yi Dou, Zhaopeng Tu, Xing Wang, Shuming Shi, and Tong Zhang.
\newblock Exploiting deep representations for neural machine translation.
\newblock In {\em Conference on Empirical Methods in Natural Language
  Processing (EMNLP)}, 2018.

\bibitem{gan2020large}
Zhe Gan, Yen-Chun Chen, Linjie Li, Chen Zhu, Yu Cheng, and Jingjing Liu.
\newblock Large-scale adversarial training for vision-and-language
  representation learning.
\newblock In {\em Conference on Neural Information Processing Systems
  (NeurIPS)}, 2020.

\bibitem{nce1}
Michael Gutmann and Aapo Hyv{\"a}rinen.
\newblock Noise-contrastive estimation: A new estimation principle for
  unnormalized statistical models.
\newblock In {\em International Conference on Artificial Intelligence and
  Statistics (AISTATS)}, 2010.

\bibitem{he2020deberta}
Pengcheng He, Xiaodong Liu, Jianfeng Gao, and Weizhu Chen.
\newblock De{BERT}a: Decoding-enhanced bert with disentangled attention.
\newblock {\em arXiv preprint}, 2020.

\bibitem{hendricks2021decoupling}
Lisa~Anne Hendricks, John Mellor, Rosalia Schneider, Jean-Baptiste Alayrac, and
  Aida Nematzadeh.
\newblock Decoupling the role of data, attention, and losses in multimodal
  transformers.
\newblock {\em Transactions of the Association for Computational Linguistics
  (TACL)}, 2021.

\bibitem{huang2017densely}
Gao Huang, Zhuang Liu, Laurens Van Der~Maaten, and Kilian~Q Weinberger.
\newblock Densely connected convolutional networks.
\newblock In {\em Conference on Computer Vision and Pattern Recognition
  (CVPR)}, 2017.

\bibitem{huang2021seeing}
Zhicheng Huang, Zhaoyang Zeng, Yupan Huang, Bei Liu, Dongmei Fu, and Jianlong
  Fu.
\newblock Seeing out of the box: End-to-end pre-training for vision-language
  representation learning.
\newblock In {\em Conference on Computer Vision and Pattern Recognition
  (CVPR)}, 2021.

\bibitem{huang2020pixel}
Zhicheng Huang, Zhaoyang Zeng, Bei Liu, Dongmei Fu, and Jianlong Fu.
\newblock Pixel-{BERT}: Aligning image pixels with text by deep multi-modal
  transformers.
\newblock {\em arXiv preprint}, 2020.

\bibitem{jawahar2019does}
Ganesh Jawahar, Beno{\^\i}t Sagot, and Djam{\'e} Seddah.
\newblock What does {BERT} learn about the structure of language?
\newblock In {\em Annual Meeting of the Association for Computational
  Linguistics (ACL)}, 2019.

\bibitem{jia2021scaling}
Chao Jia, Yinfei Yang, Ye Xia, Yi-Ting Chen, Zarana Parekh, Hieu Pham, Quoc~V
  Le, Yunhsuan Sung, Zhen Li, and Tom Duerig.
\newblock Scaling up visual and vision-language representation learning with
  noisy text supervision.
\newblock {\em arXiv preprint}, 2021.

\bibitem{nce2}
Rafal Jozefowicz, Oriol Vinyals, Mike Schuster, Noam Shazeer, and Yonghui Wu.
\newblock Exploring the limits of language modeling.
\newblock {\em arXiv preprint}, 2016.

\bibitem{kamath2021mdetr}
Aishwarya Kamath, Mannat Singh, Yann LeCun, Gabriel Synnaeve, Ishan Misra, and
  Nicolas Carion.
\newblock Mdetr-modulated detection for end-to-end multi-modal understanding.
\newblock In {\em International Conference on Computer Vision (ICCV)}, 2021.

\bibitem{kim2021vilt}
Wonjae Kim, Bokyung Son, and Ildoo Kim.
\newblock {ViLT}: Vision-and-language transformer without convolution or region
  supervision.
\newblock In {\em International Conference on Machine Learning (ICML)}, 2021.

\bibitem{krishna2016visual}
Ranjay Krishna, Yuke Zhu, Oliver Groth, Justin Johnson, Kenji Hata, Joshua
  Kravitz, Stephanie Chen, Yannis Kalantidis, Li-Jia Li, David~A Shamma, et~al.
\newblock Visual genome: Connecting language and vision using crowdsourced
  dense image annotations.
\newblock {\em International Journal of Computer Vision (IJCV)}, 2017.

\bibitem{krizhevsky2009learning}
Alex Krizhevsky, Geoffrey Hinton, et~al.
\newblock Learning multiple layers of features from tiny images.
\newblock 2009.

\bibitem{lan2019albert}
Zhenzhong Lan, Mingda Chen, Sebastian Goodman, Kevin Gimpel, Piyush Sharma, and
  Radu Soricut.
\newblock Albert: A lite bert for self-supervised learning of language
  representations.
\newblock In {\em International Conference on Learning Representations (ICLR)},
  2020.

\bibitem{li2021align}
Junnan Li, Ramprasaath~R Selvaraju, Akhilesh~Deepak Gotmare, Shafiq Joty,
  Caiming Xiong, and Steven Hoi.
\newblock Align before fuse: Vision and language representation learning with
  momentum distillation.
\newblock In {\em Conference on Neural Information Processing Systems
  (NeurIPS)}, 2021.

\bibitem{li2019visualbert}
Liunian~Harold Li, Mark Yatskar, Da Yin, Cho-Jui Hsieh, and Kai-Wei Chang.
\newblock Visual{BERT}: A simple and performant baseline for vision and
  language.
\newblock {\em arXiv preprint}, 2019.

\bibitem{li2020unimo}
Wei Li, Can Gao, Guocheng Niu, Xinyan Xiao, Hao Liu, Jiachen Liu, Hua Wu, and
  Haifeng Wang.
\newblock Unimo: Towards unified-modal understanding and generation via
  cross-modal contrastive learning.
\newblock In {\em Annual Meeting of the Association for Computational
  Linguistics (ACL)}, 2021.

\bibitem{li2020oscar}
Xiujun Li, Xi Yin, Chunyuan Li, Pengchuan Zhang, Xiaowei Hu, Lei Zhang, Lijuan
  Wang, Houdong Hu, Li Dong, Furu Wei, et~al.
\newblock Oscar: Object-semantics aligned pre-training for vision-language
  tasks.
\newblock In {\em European Conference on Computer Vision (ECCV)}, 2020.

\bibitem{lin2014microsoft}
Tsung-Yi Lin, Michael Maire, Serge Belongie, James Hays, Pietro Perona, Deva
  Ramanan, Piotr Doll{\'a}r, and C~Lawrence Zitnick.
\newblock Microsoft {COCO}: Common objects in context.
\newblock In {\em European Conference on Computer Vision (ECCV)}, 2014.

\bibitem{liu2019text}
Yang Liu and Mirella Lapata.
\newblock Text summarization with pretrained encoders.
\newblock In {\em Conference on Empirical Methods in Natural Language
  Processing (EMNLP)}, 2019.

\bibitem{liu2019roberta}
Yinhan Liu, Myle Ott, Naman Goyal, Jingfei Du, Mandar Joshi, Danqi Chen, Omer
  Levy, Mike Lewis, Luke Zettlemoyer, and Veselin Stoyanov.
\newblock Ro{BERT}a: A robustly optimized bert pretraining approach.
\newblock {\em arXiv preprint}, 2019.

\bibitem{liu2021swin}
Ze Liu, Yutong Lin, Yue Cao, Han Hu, Yixuan Wei, Zheng Zhang, Stephen Lin, and
  Baining Guo.
\newblock Swin transformer: Hierarchical vision transformer using shifted
  windows.
\newblock In {\em International Conference on Computer Vision (ICCV)}, 2021.

\bibitem{loshchilov2018decoupled}
Ilya Loshchilov and Frank Hutter.
\newblock Decoupled weight decay regularization.
\newblock In {\em International Conference on Learning Representations (ICLR)},
  2018.

\bibitem{lu2019vilbert}
Jiasen Lu, Dhruv Batra, Devi Parikh, and Stefan Lee.
\newblock Vilbert: Pretraining task-agnostic visiolinguistic representations
  for vision-and-language tasks.
\newblock In {\em Conference on Neural Information Processing Systems
  (NeurIPS)}, 2019.

\bibitem{ordonez2011im2text}
Vicente Ordonez, Girish Kulkarni, and Tamara Berg.
\newblock Im2text: Describing images using 1 million captioned photographs.
\newblock In {\em Conference on Neural Information Processing Systems
  (NeurIPS)}, 2011.

\bibitem{plummer2015flickr30k}
Bryan~A Plummer, Liwei Wang, Chris~M Cervantes, Juan~C Caicedo, Julia
  Hockenmaier, and Svetlana Lazebnik.
\newblock Flickr30k {E}ntities: Collecting region-to-phrase correspondences for
  richer image-to-sentence models.
\newblock In {\em International Conference on Computer Vision (ICCV)}, 2015.

\bibitem{radford2021learning}
Alec Radford, Jong~Wook Kim, Chris Hallacy, Aditya Ramesh, Gabriel Goh,
  Sandhini Agarwal, Girish Sastry, Amanda Askell, Pamela Mishkin, Jack Clark,
  et~al.
\newblock Learning transferable visual models from natural language
  supervision.
\newblock In {\em International Conference on Machine Learning (ICML)}, 2021.

\bibitem{raffel2020exploring}
Colin Raffel, Noam Shazeer, Adam Roberts, Katherine Lee, Sharan Narang, Michael
  Matena, Yanqi Zhou, Wei Li, and Peter~J Liu.
\newblock Exploring the limits of transfer learning with a unified text-to-text
  transformer.
\newblock {\em Journal of Machine Learning Research (JMLR)}, 2020.

\bibitem{ramesh2021zero}
Aditya Ramesh, Mikhail Pavlov, Gabriel Goh, Scott Gray, Chelsea Voss, Alec
  Radford, Mark Chen, and Ilya Sutskever.
\newblock Zero-shot text-to-image generation.
\newblock In {\em International Conference on Machine Learning (ICML)}, 2021.

\bibitem{ren2015faster}
Shaoqing Ren, Kaiming He, Ross Girshick, and Jian Sun.
\newblock Faster {R-CNN}: Towards real-time object detection with region
  proposal networks.
\newblock {\em IEEE Transactions on Pattern Analysis and Machine Intelligence
  (TPAMI)}, 2016.

\bibitem{selvaraju2017grad}
Ramprasaath~R Selvaraju, Michael Cogswell, Abhishek Das, Ramakrishna Vedantam,
  Devi Parikh, and Dhruv Batra.
\newblock Grad-{CAM}: Visual explanations from deep networks via gradient-based
  localization.
\newblock In {\em International Conference on Computer Vision (ICCV)}, 2017.

\bibitem{sennrich2016neural}
Rico Sennrich, Barry Haddow, and Alexandra Birch.
\newblock Neural machine translation of rare words with subword units.
\newblock In {\em Annual Meeting of the Association for Computational
  Linguistics (ACL)}, 2016.

\bibitem{sharma2018conceptual}
Piyush Sharma, Nan Ding, Sebastian Goodman, and Radu Soricut.
\newblock Conceptual captions: A cleaned, hypernymed, image alt-text dataset
  for automatic image captioning.
\newblock In {\em Annual Meeting of the Association for Computational
  Linguistics (ACL)}, 2018.

\bibitem{shen2021much}
Sheng Shen, Liunian~Harold Li, Hao Tan, Mohit Bansal, Anna Rohrbach, Kai-Wei
  Chang, Zhewei Yao, and Kurt Keutzer.
\newblock How much can clip benefit vision-and-language tasks?
\newblock {\em arXiv preprint}, 2021.

\bibitem{su2019vl}
Weijie Su, Xizhou Zhu, Yue Cao, Bin Li, Lewei Lu, Furu Wei, and Jifeng Dai.
\newblock {VL-BERT}: Pre-training of generic visual-linguistic representations.
\newblock In {\em International Conference on Learning Representations (ICLR)},
  2019.

\bibitem{suhr2018corpus}
Alane Suhr, Stephanie Zhou, Ally Zhang, Iris Zhang, Huajun Bai, and Yoav Artzi.
\newblock A corpus for reasoning about natural language grounded in
  photographs.
\newblock In {\em Annual Meeting of the Association for Computational
  Linguistics (ACL)}, 2019.

\bibitem{tan-bansal-2019-lxmert}
Hao Tan and Mohit Bansal.
\newblock {LXMERT}: Learning cross-modality encoder representations from
  transformers.
\newblock In {\em Conference on Empirical Methods in Natural Language
  Processing (EMNLP)}, 2019.

\bibitem{touvron2020deit}
Hugo Touvron, Matthieu Cord, Matthijs Douze, Francisco Massa, Alexandre
  Sablayrolles, and Herv\'e J\'egou.
\newblock Training data-efficient image transformers \& distillation through
  attention.
\newblock {\em arXiv preprint}, 2020.

\bibitem{touvron2021going}
Hugo Touvron, Matthieu Cord, Alexandre Sablayrolles, Gabriel Synnaeve, and
  Herv{\'e} J{\'e}gou.
\newblock Going deeper with image transformers.
\newblock {\em arXiv preprint}, 2021.

\bibitem{van2017neural}
Aaron van~den Oord, Oriol Vinyals, and Koray Kavukcuoglu.
\newblock Neural discrete representation learning.
\newblock In {\em Conference on Neural Information Processing Systems
  (NeurIPS)}, 2017.

\bibitem{vaswani2017attention}
Ashish Vaswani, Noam Shazeer, Niki Parmar, Jakob Uszkoreit, Llion Jones,
  Aidan~N Gomez, {\L}ukasz Kaiser, and Illia Polosukhin.
\newblock Attention is all you need.
\newblock In {\em Conference on Neural Information Processing Systems
  (NeurIPS)}, 2017.

\bibitem{wang2018glue}
Alex Wang, Amanpreet Singh, Julian Michael, Felix Hill, Omer Levy, and Samuel
  Bowman.
\newblock {GLUE}: A multi-task benchmark and analysis platform for natural
  language understanding.
\newblock In {\em International Conference on Learning Representations (ICLR)},
  2019.

\bibitem{wang2020gradient}
Zirui Wang, Yulia Tsvetkov, Orhan Firat, and Yuan Cao.
\newblock Gradient vaccine: Investigating and improving multi-task optimization
  in massively multilingual models.
\newblock In {\em International Conference on Learning Representations (ICLR)},
  2020.

\bibitem{wang2021simvlm}
Zirui Wang, Jiahui Yu, Adams~Wei Yu, Zihang Dai, Yulia Tsvetkov, and Yuan Cao.
\newblock Simvlm: Simple visual language model pretraining with weak
  supervision.
\newblock {\em arXiv preprint}, 2021.

\bibitem{xie2019visual}
Ning Xie, Farley Lai, Derek Doran, and Asim Kadav.
\newblock Visual entailment: A novel task for fine-grained image understanding.
\newblock {\em arXiv preprint}, 2019.

\bibitem{xue2021probing}
Hongwei Xue, Yupan Huang, Bei Liu, Houwen Peng, Jianlong Fu, Houqiang Li, and
  Jiebo Luo.
\newblock Probing inter-modality: Visual parsing with self-attention for
  vision-language pre-training.
\newblock In {\em Conference on Neural Information Processing Systems
  (NeurIPS)}, 2021.

\bibitem{yu2018deep}
Fisher Yu, Dequan Wang, Evan Shelhamer, and Trevor Darrell.
\newblock Deep layer aggregation.
\newblock In {\em Conference on Computer Vision and Pattern Recognition
  (CVPR)}, 2018.

\bibitem{yu2020gradient}
Tianhe Yu, Saurabh Kumar, Abhishek Gupta, Sergey Levine, Karol Hausman, and
  Chelsea Finn.
\newblock Gradient surgery for multi-task learning.
\newblock In {\em Conference on Neural Information Processing Systems
  (NeurIPS)}, 2020.

\bibitem{yuan2021florence}
Lu Yuan, Dongdong Chen, Yi-Ling Chen, Noel Codella, Xiyang Dai, Jianfeng Gao,
  Houdong Hu, Xuedong Huang, Boxin Li, Chunyuan Li, Ce Liu, Mengchen Liu,
  Zicheng Liu, Yumao Lu, Yu Shi, Lijuan Wang, Jianfeng Wang, Bin Xiao, Zhen
  Xiao, Jianwei Yang, Michael Zeng, Luowei Zhou, and Pengchuan Zhang.
\newblock Florence: A new foundation model for computer vision.
\newblock {\em arXiv preprint}, 2021.

\bibitem{yuan2021volo}
Li Yuan, Qibin Hou, Zihang Jiang, Jiashi Feng, and Shuicheng Yan.
\newblock Volo: Vision outlooker for visual recognition.
\newblock {\em arXiv preprint}, 2021.

\bibitem{zhang2021vinvl}
Pengchuan Zhang, Xiujun Li, Xiaowei Hu, Jianwei Yang, Lei Zhang, Lijuan Wang,
  Yejin Choi, and Jianfeng Gao.
\newblock {VinVL}: Revisiting visual representations in vision-language models.
\newblock In {\em Conference on Computer Vision and Pattern Recognition
  (CVPR)}, 2021.

\end{thebibliography}
}

\clearpage
\appendix
\section{Implementation Details}
\paragraph{Datasets.} The statistics of our pre-training datasets is shown in Table~\ref{tab:stats}. Following many previous work~\cite{chen2020uniter,kim2021vilt,li2021align}, we pre-train the models with four datasets, including COCO, Visual Genome, Conceptual Captions and SBU Captions, consisting of about 4M images and 9M image-caption pairs in total. 

For the downstream tasks, we test the models on VQAv2~\cite{antol2015vqa} for visual question answering, NLVR$^2$~\cite{suhr2018corpus} for visual reasoning, COCO~\cite{lin2014microsoft} and Flickr30k~\cite{plummer2015flickr30k} for image-text retrieval, and SNLI-VE~\cite{xie2019visual} for visual entailment.  We use the standard splits for all the datasets except for VQAv2, where we follow standard practice~\cite{chen2020uniter,kim2021vilt,li2021align} to train the models with both its training and development data, and treat its test-dev set as the development set. Note that we do not use the Visual Genome VQA data for data augmentation in our VQA settings.

\begin{table}[!t]
\tablestyle{5pt}{1.0}
\def\w{25pt} 
\scalebox{1.0}{
  \begin{tabular}{c|cccc}
    & \bf COCO & \bf VG & \bf CC & \bf SBU \\
    \shline
\#Images & 113K & 108K & 3.1M & 875K \\
\#Captions & 567K & 5.4M & 3.1M & 875K \\
  \end{tabular}
  }
  \caption{Statistics of the pre-training datasets.}
  \label{tab:stats}
  \vspace{-3mm}
\end{table}

\paragraph{Pre-training Settings.} We pre-train our best models using the AdamW optimizer~\cite{loshchilov2018decoupled} with the learning rates set to 1e-5 for the bottom image and text encoders and 5e-5 for the cross-modal module. The warm-up ratio is set to 10\%, and the learning rate is linearly decayed to 0 after 10\% of the total training steps. The batch size, hidden size, and number of heads are set to 4096, 768, 12, respectively. We pre-train the models for 100k steps on 8 NVIDIA A100 GPUs, which takes around 3 days for \ModelName-CLIP-ViT$_{\text{BASE}-32}$ and 8 days for \ModelName-Swin$_{\text{BASE}}$ and \ModelName-CLIP-ViT$_{\text{BASE}-16}$.

\paragraph{Fine-tuning Settings.} For the downstream tasks, we perform grid searches over the learning rates and image resolutions. The learning rates and image resolutions are selected from \{1e-6, 2e-6, 5e-6, 1e-5\} and \{288, 384, 576\}, respectively. We apply RandAugment~\cite{cubuk2020randaugment} during finetuning following previous work~\cite{kim2021vilt,li2021align}.

\section{Inference Time}
We measure the inference time of different models as in Table~\ref{tab:infer_time}. First, as shown in~\cite{kim2021vilt}, their ViT-based model is much faster than previous region-feature-based VLP models. In our setting, we measure the average inference time of processing 1 VQA instance on 1 NVIDIA V100 GPU. We find that while our model can be slower than the ViLT model, it is still significantly faster than region-feature-based models and comparable to other ViT-based ones. In addition, we can achieve much stronger performance on downstream tasks than other models. 

\begin{table}[!t]
\tablestyle{5pt}{1.0}
\def\w{25pt} 
\scalebox{1.0}{
  \begin{tabular}{l|cc|c}
   \bf Model & \bf Time (~\cite{kim2021vilt}) & \bf Time (ours) & \bf VQAv2 \\ 
    \shline
      ViLBERT & 920 & - & 70.55 \\
      VisualBERT & 925 & - & 70.80\\
      LXMERT & 900 & -  & 72.42\\
      UNITER-Base & 900 & - & 72.70\\
      OSCAR-Base & 900 & - & 73.16\\
      VinVL-Base & 650 & - & 75.95\\
      PixelBERT-X152 & 160 & - & 74.45\\
      CLIP-ViL (ResNet50x4) & - & 57 & 76.70 \\ 
      ViLT & 15 & 26  & 71.26\\
      ALBEF (14M) & - & 52 & 76.04 \\
      \cdashline{1-4}
      \ModelName-Swin$_{\text{BASE}}$  & - &  59 & 76.42 \\
      \ModelName-CLIP-ViT$_{\text{BASE}}$  & - & 53 & 77.64 \\
  \end{tabular}
  }
  \caption{Inference time (ms) of different models. We report the inference time measured by~\cite{kim2021vilt} and in our setting. We also list the model performance on the VQAv2 test-std set.}
  \label{tab:infer_time}
  \vspace{-3mm}
\end{table}

\section{Image Captioning}
While in this paper we mainly focus on finetuning our models for discriminative downstream tasks such as visual question answering, here we investigate if our models can also be applied to generative tasks. Specifically, we finetune our models on the COCO image captioning task.

We finetune our \ModelName-CLIP-ViT$_{\text{BASE}}$ model for 5 epochs using the standard maximum likelihood estimation objective. At each decoding step, instead of using the causal attention mechanism, the input image and all the text tokens can attend to all the generated text tokens so as to minimize the discrepancy between pre-training and finetuning. We use beam search with the beam size set to 5.

As shown in Table~\ref{tab:caption}, we can achieve reasonable performance on image captioning even though our model employs an encoder-only architecture. We expect that an encoder-decoder model would be more suitable for generative tasks, which we leave as future work.

\begin{table}[!t]
\tablestyle{5pt}{1.0}
\def\w{20pt} 
\scalebox{0.95}{
  \begin{tabular}{l|ccccccc}
  \bf Model  (\#Pre-training Images)& \bf BLEU & \bf METEOR & \bf Cider & \bf SPICE\\
    \shline
    OSCAR$_{\text{BASE}}$ (4M) & 36.5 & 30.3 & 123.7 & 23.1 \\
    VinVL$_{\text{BASE}}$  (5.6M) &38.2 & 30.3 & 129.3 & 23.6\\
    SimVLM$_{\text{BASE}}$ (1.8B) & 39.0 & 32.9 & 134.8 & 24.0 \\
         \cdashline{1-5}
    \ModelName-CLIP-ViT$_{\text{BASE}}$ (4M) & 38.8 & 30.0 & 128.2 & 23.0 \\
  \end{tabular}
  }
  \caption{Image captioning results of different models trained with maximum likelihood estimation on COCO.}
  \label{tab:caption}
\end{table}

\section{Multi-scale Feature Fusion}
For the pre-trained text and visual encoders, different layers can contain different types of information. For example, \cite{jawahar2019does} finds that the intermediate
layers of BERT encode a rich hierarchy of linguistic information, with surface features at the bottom,
syntactic features in the middle and semantic
features at the top. Aggregating the features at different layers has demonstrated to be helpful in both vision~\cite{huang2017densely,yu2018deep} and language~\cite{bapna2018training,dou2018exploiting}. Therefore, in this part, we investigate if we can use feature fusion techniques to better utilize the information embedded at different layers of the pre-trained encoders.

\paragraph{Method.} Based on some preliminary explorations, here we adopt a simple fusion strategy and only fuse the representations of the text and image encoders but not the cross-modal layers on the top. Specifically, given a text token or image patch $x_i$, we first feed it into a text or image encoder on the bottom of our model (\eg, BERT), and get its representations $\{ h(x_i^j) \}_{j=0}^N$ at different layers, where $N$ is the number of layers of the encoder. Then, we compute a gate value for each layer and perform a weighted sum to get the final representation of $x_i$:
\begin{equation}
    o(x_i) = h(x_i^N) + \sum_{j=0}^{N-1} g(h(x_i^j)) h(x_i^j),
\end{equation}
where $g$ is a linear transformation function. We then feed $o(x_i)$ to the top cross-modal layers. Note that the fusion can be done in both the text and visual encoders.

\begin{table}[!t]
\tablestyle{5pt}{1.0}
\def\w{20pt} 
\scalebox{1.0}{
  \begin{tabular}{l|ccc}
    \multirow{2}{*}{\bf Model}  & \bf VQAv2 &  \multicolumn{2}{c}{\bf Flickr-ZS} \\
     & \bf test-dev & \bf IR & \bf TR \\
    \shline
    \emph{w/o pre-training} \\
    \hline
    \ModelName-Swin$_{\text{BASE}}$ w/o fusion & 72.38 & - & -\\
    \ModelName-Swin$_{\text{BASE}}$ w/ fusion & 72.91 &- & -\\
     \cdashline{1-4}
    \ModelName-CLIP-ViT$_{\text{BASE}}$ w/o fusion & 71.75 & - & -\\
     \ModelName-CLIP-ViT$_{\text{BASE}}$ w/ fusion & 72.92 &- & -\\
    \hline
    \emph{with pre-training} \\
    \hline
    \ModelName-Swin$_{\text{BASE}}$ w/o fusion& 76.43 & 71.68 & 85.30\\
    \ModelName-Swin$_{\text{BASE}}$ w/ fusion & 76.31 & 70.58 & 83.70\\
     \cdashline{1-4}
     \ModelName-CLIP-ViT$_{\text{BASE}}$ w/o fusion & 77.19 & 76.64 & 89.60 \\
    \ModelName-CLIP-ViT$_{\text{BASE}}$ w/ fusion & 77.06 & 76.26 & 88.00\\
  \end{tabular}
  }
  \caption{The fusion strategy improves the model performance without vision-and-language pre-training but can degrade the model performance after pre-training.}
  \label{tab:fusion}
\end{table}
\paragraph{Results.} We pre-train the models using the co-attention model with RoBERTa as the text encoder and Swin Transformer as the visual encoder. We evaluate the models both with and without VLP following the default settings. Because Swin Transformers have different numbers of image representations at different layers, we perform an average pooling so that each layer has 12$\times$12 patch representations. As shown in~\cref{tab:fusion}, while the fusion strategy can improve the model performance by a small margin without VLP, it can degrade the model performance after pre-training. We hypothesize that this is because after the pre-training, the VLP model can learn how to well utilize the representations in the pre-trained encoders and layer fusion is not necessary.

\section{Correlation between \textit{Vision-and-Language} Tasks and \textit{Vision} or \textit{Language} Tasks}
\begin{figure}[t!]
  \centering
  \begin{subfigure}{0.495\linewidth}
    \includegraphics[width=1.0\linewidth]{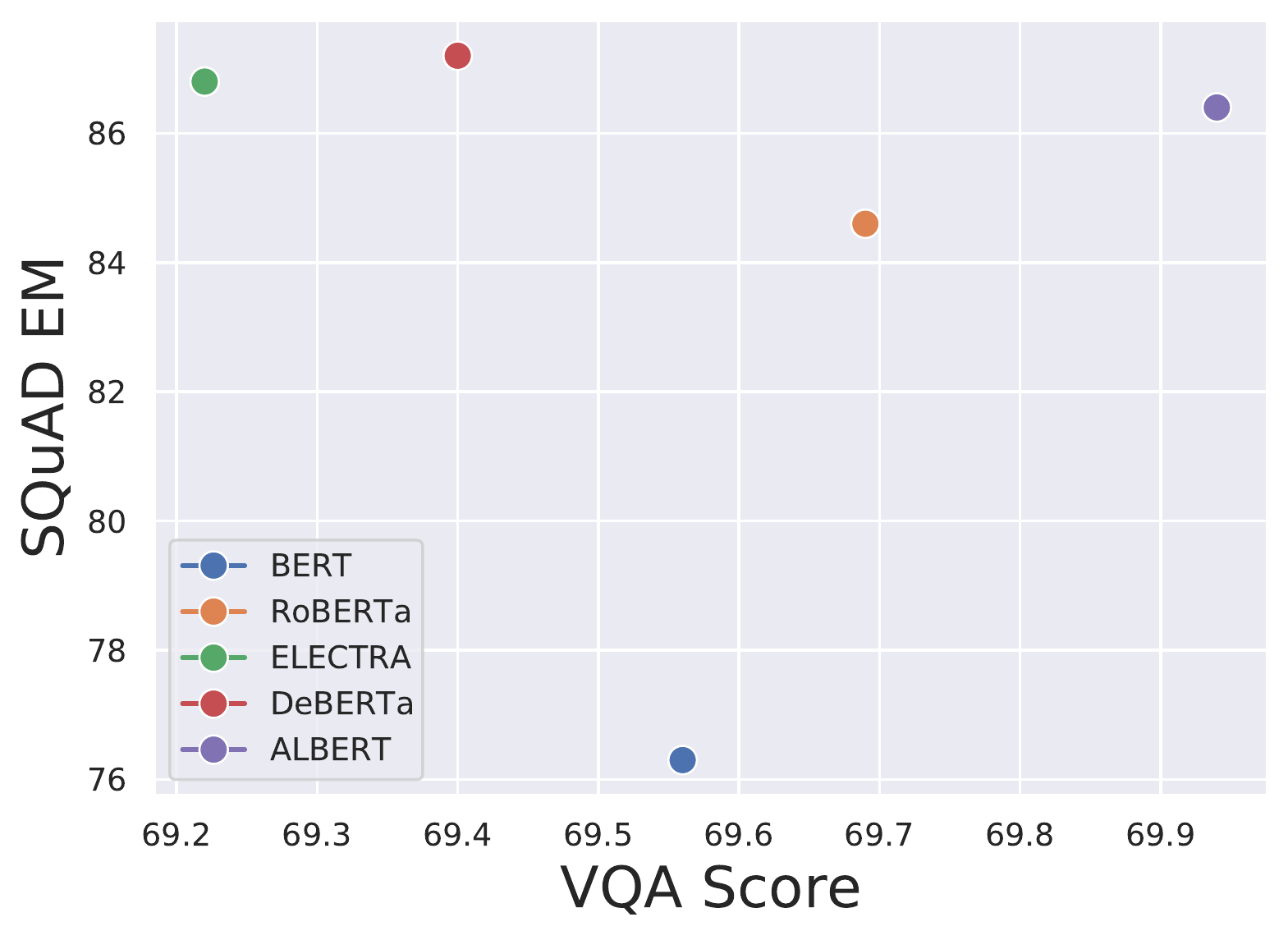}
    \caption{Vision-and-Language vs. Language}
    \label{fig:cor-text}
  \end{subfigure}
  \begin{subfigure}{0.495\linewidth}
    \includegraphics[width=1.0\linewidth]{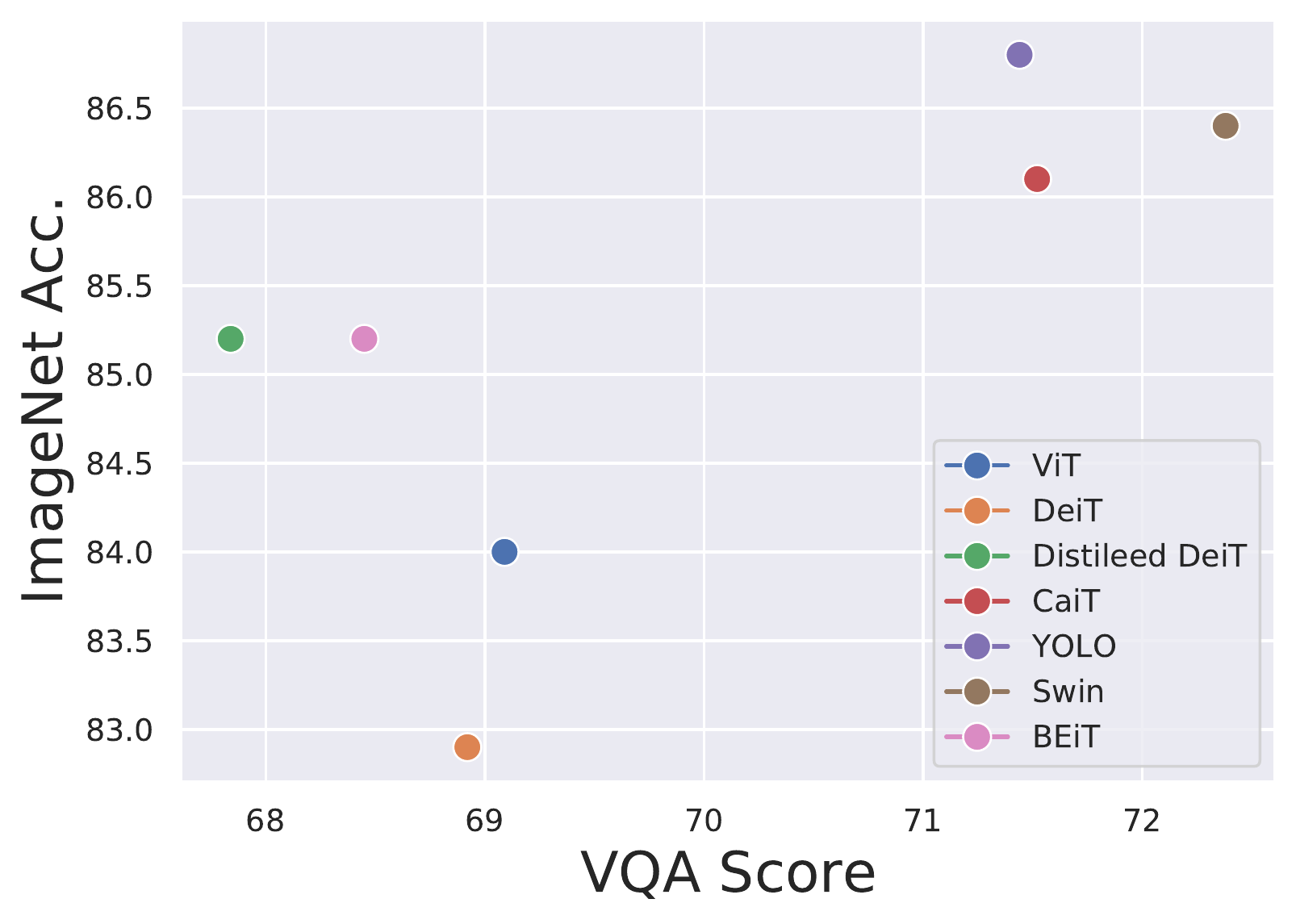}
    \caption{Vision-and-Language vs. Vision}
    \label{fig:cor-vision}
  \end{subfigure}
  \caption{Correlation between model performance on \textit{vision-and-language} tasks and pure \textit{vision} or \textit{language} tasks.}
  \label{fig:cor}
  \vspace{-3mm}
\end{figure}

In this section, we perform a quantitative analysis of the correlation between model performance on \textit{vision-and-language} tasks and pure \textit{vision} or \textit{language} tasks. We vary different text and vision encoders, and plot the model performance on the VQAv2 test-dev set and SQuAD or ImageNet datasets as in Figure~\ref{fig:cor}. We also compute the Pearson correlations in both cases. We find that the Pearson correlation coefficients and p-values are -0.09 and 0.88 in the VL vs. L setting, and 0.41 and 0.36 in the VL vs. V setting, indicating that there exists little to none correlations between the model performance on VL tasks and V or L tasks.

\begin{table*}[!t]
\tablestyle{2pt}{1.0}
\def\w{20pt} 
\scalebox{1.0}{
  \begin{tabular}{c|ccccccccccc}
    \multirow{1}{*}{\bf Text Encoder} & \bf QQP & \bf  MNLI & \bf QNLI & \bf SST2 & \bf CoLA & \bf MRPC & \bf STSB & \bf RTE  \\
    \shline
  Before VLP & 91.31$\pm$0.15 & 87.53$\pm$0.24 & 92.61$\pm$0.34 & 94.38$\pm$0.20& 58.72$\pm$0.73 & 91.03$\pm$0.59 & 90.15$\pm$0.18 & 71.24$\pm$3.07 \\
  After VLP  & 91.34$\pm$0.08 & 87.38$\pm$0.18 & 92.67$\pm$0.06 & 93.92$\pm$0.50 & 57.88$\pm$0.79 & 90.57$\pm$0.78 & 89.93$\pm$0.46 & 70.28$\pm$2.00   \\
  \end{tabular}
  }
  \caption{Performance of text encoders (RoBERTa-base) on GLUE dev sets before and after VLP. The image encoder during VLP is CLIP-ViT-224/16. We report average scores and standard deviations over three runs of different random seeds.}
  \label{tab:text-only}
\end{table*}

\begin{table}[!t]
\tablestyle{5pt}{1.0}
\def\w{20pt} 
\scalebox{1.0}{
  \begin{tabular}{c|ccccc}
    \multirow{2}{*}{\bf Image Encoder} &  \multicolumn{2}{c}{\bf Before VLP} &  \multicolumn{2}{c}{\bf After VLP} \\
     & \bf CF10 & \bf CF100  & \bf CF10  & \bf CF100 \\
    \shline
  Swin-Base-384/32 & 97.00 & 89.15 & 97.99 & 90.26\\
  CLIP-ViT-224/16  & 95.85 & 82.60 & 94.92 & 81.90\\
  \end{tabular}
  }
  \caption{Linear probe performance on CIFAR-10 and CIFAR-100. The text encoder during VLP is RoBERTa-base.}
  \label{tab:image-only}
\end{table}

\section{Unimodal Tasks}
We also investigate the model performance on unimodal tasks after VLP. For text-only tasks, we finetune the bottom text encoders on GLUE tasks; for image-only tasks, we fit a linear classifier on the learned representations of image encoders on CIFAR-10 and CIFAR-100~\cite{krizhevsky2009learning}.
    
We report results in Table~\ref{tab:text-only} and~\ref{tab:image-only}. As shown in the tables, our text encoder gets slightly worse performance on the GLUE tasks on average;
for image-only tasks, VLP seems to improve the model performance for Swin Transformer but not for CLIP-ViT, possibly because of domain issues. Note that in both sets of the experiments, we only use our text or image encoder and discard the rest of the networks, and how to utilize multi-modal encoder to improve uni-modal performance is an open problem and we leave it as a future direction.

\begin{table}[!t]
\tablestyle{5pt}{1.0}
\def\w{20pt} 
\scalebox{1.0}{
  \begin{tabular}{c|ccc}
    \multirow{2}{*}{\bf Pre-training Datasets} &  \multirow{2}{*}{\bf VQAv2} &  \multicolumn{2}{c}{\bf Flickr-ZS} \\
     & & \bf IR & \bf TR \\
    \shline
  COCO & 72.95 & 46.38 & 60.20  \\
CC& 73.05 & 39.84 & 55.50  \\
   SBU & 70.14 & 21.52 & 35.90  \\
    VG & 73.54 & 39.24 & 49.30\\
    \shline
  COCO+CC+SBU+VG&  74.98 &  66.08 & 78.10 \\
  \end{tabular}
  }
  \caption{Results of models pre-trained with different datasets.}
  \label{tab:pretrain-datasets}
\end{table}


\section{Analysis on Pre-training Datasets}
We also perform analysis on our pre-training datasets.
We pre-train our model on each of the pre-training datasets. We choose CLIP-ViT-224/32 as the image encoder and BERT-base-uncased as the text encoder, and employ the co-attention fusion module. We pre-train the model for 50k steps on each dataset and report the evaluation results on VQAv2 and Flickr30k zero-shot retrieval tasks.

As we can see from Table~\ref{tab:pretrain-datasets}, both data size and domain similarity contribute to the downstream task performance. CC and VG are the largest datasets and COCO most matches the downstream task domains, thus models pre-trained on the three datasets obtain the highest scores.

\section{Visualization}
\begin{figure*}
  \centering
    \includegraphics[width=1.0\linewidth]{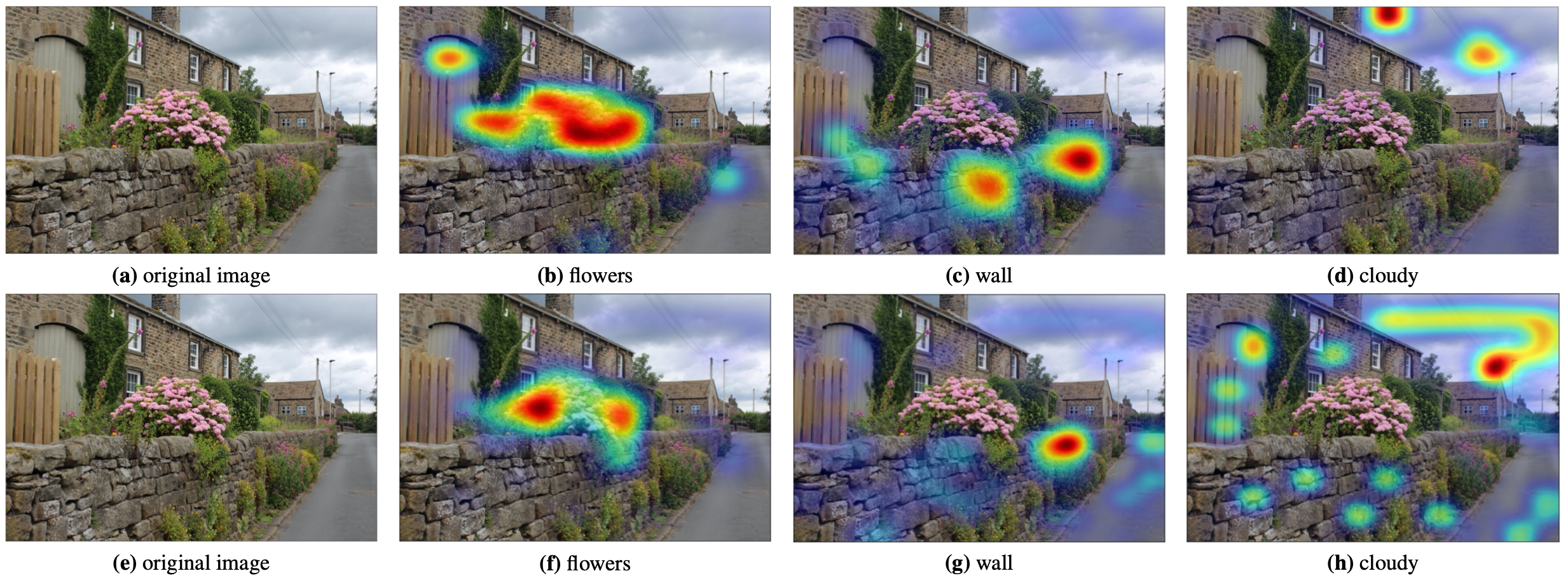}
  \caption{Visualization of the attention maps of text tokens in the caption ``a display of \textit{flowers} growing out and over the retaining \textit{wall} in front of cottages on a \textit{cloudy} day.'' The first and second rows correspond to the results of \ModelName-Swin$_{\text{BASE}}$ and \ModelName-CLIP-ViT$_{\text{BASE}}$, respectively.}
  \label{fig:vis}
\end{figure*}

\begin{figure*}
  \centering
   \includegraphics[width=1.0\linewidth]{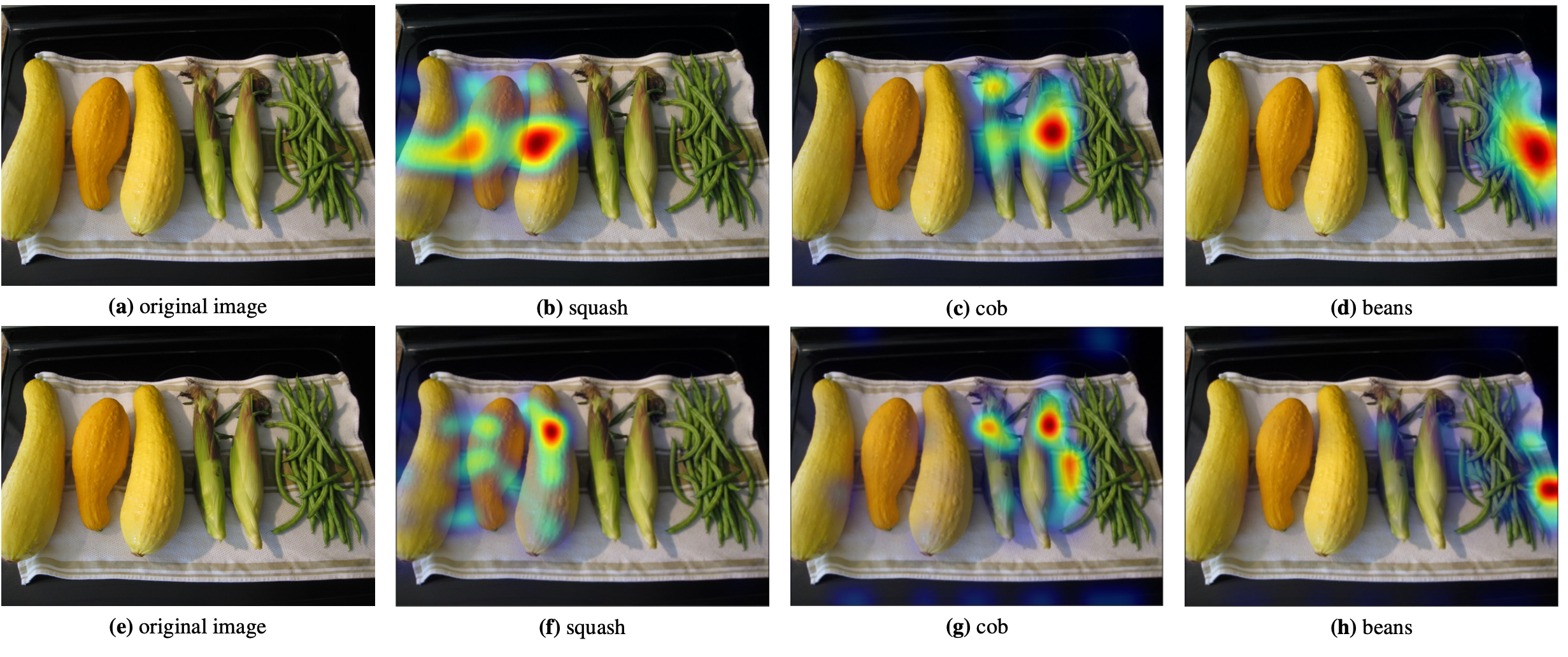}
  \caption{Visualization of the attention maps of text tokens in the caption ``yellow \textit{squash}, corn on the \textit{cob} and green \textit{beans} laid out on a white cloth.'' The first and second rows correspond to the results of \ModelName-Swin$_{\text{BASE}}$ and \ModelName-CLIP-ViT$_{\text{BASE}}$, respectively.
}
  \label{fig:vis2}
\end{figure*}

In this section, we use Grad-CAM~\cite{selvaraju2017grad} to visualize our models. Specifically, we visualize the cross-attention maps of the pre-trained models corresponding to individual words when performing masked language modeling. As shown in Figure~\ref{fig:vis} and~\ref{fig:vis2}, both our Swin Transformer-based and CLIP-ViT-based models can correctly attend to the corresponding regions given different words, suggesting that our models can learn visual grounding implicitly during pre-training. 

\section{Limitations}

While we have demonstrated the effectiveness of our models across different tasks, our models still have several limitations:
\paragraph{Generative Tasks.} We mainly focus on discriminative tasks such as visual question answering and visual reasoning in this paper, while generative tasks such as image captioning are under-investigated. We perform experiments on the COCO image captioning data in Appendix, and will investigate more on this in the future.
\paragraph{Scalability.} In our current settings, we pre-train the models with 4M or 14M images, thus it is unclear how the model performance would be if we pre-train the models with larger datasets and we are actively experimenting in this direction.
\paragraph{English Data.} So far, we only experiment on the English data, and it is worth investigating if our models can generalize to other languages as well, which we leave as a future direction.

\end{document}